\documentclass[journal]{IEEEtran}
\ifCLASSINFOpdf
\else
\fi
\usepackage{times}
\usepackage{epsfig}
\usepackage{graphicx}
\usepackage{amsmath}
\usepackage{amssymb}
\usepackage{multirow}
\usepackage{bm}
\usepackage{xcolor}
\usepackage{cite}
\usepackage{blindtext}
\usepackage{hyperref}

\begin{document}

\title{Towards Stable Co-saliency Detection and Object Co-segmentation}

\author{Bo Li~\IEEEmembership{Member,~IEEE}, Lv Tang~\IEEEmembership{Student Member,~IEEE}, Senyun Kuang~\IEEEmembership{Student Member,~IEEE}, Mofei Song and Shouhong Ding 
\thanks{

Bo Li and Shouhong Ding are with the Youtu Lab, Tencent, Shanghai 200233, China (e-mail: libraboli@tencent.com; ericshding@tencent.com).

Lv Tang is the corresponding author of this work. He was with the State Key Laboratory for Novel Software Technology, Nanjing University, Nanjing 210000, China. He is now an independent researcher and resides in Beijing 100049, China (e-mail: luckybird1994@gmail.com).

Senyun Kuang is with the School of Information Science and Technology, Southwest Jiaotong University, Chengdu 611756, China (e-mail: sykuang@my.swjtu.edu.cn).

Mofei Song is with the Key Laboratory of Computer Network and Information Integration (Ministry of Education), School of Computer Science and Engineering, Southeast University, Nanjing 210096, China (e-mail: songmf@seu.edu.cn).}}

\markboth{IEEE Transactions on Image Processing}%
{Shell \MakeLowercase{\textit{et al.}}: Bare Demo of IEEEtran.cls for IEEE Journals}



\maketitle

\begin{abstract}
In this paper, we present a novel model for simultaneous stable co-saliency detection (CoSOD) and object co-segmentation (CoSEG). To detect co-saliency (segmentation) accurately, the core problem is to well model inter-image relations between an image group. Some methods design sophisticated modules, such as recurrent neural network (RNN), to address this problem. However, order-sensitive problem is the major drawback of RNN, which heavily affects the stability of proposed CoSOD (CoSEG) model. In this paper, inspired by RNN-based model, we first propose a multi-path stable recurrent unit (MSRU), containing dummy orders mechanisms (DOM) and recurrent unit (RU). Our proposed MSRU not only helps CoSOD (CoSEG) model captures robust inter-image relations, but also reduces order-sensitivity, resulting in a more stable inference and training process. { Moreover, we design a cross-order contrastive loss (COCL) that can further address order-sensitive problem by pulling close the feature embedding generated from different input orders.} We validate our model on five widely used CoSOD datasets (CoCA, CoSOD3k, Cosal2015, iCoseg and MSRC), and three widely used datasets (Internet, iCoseg and PASCAL-VOC) for object co-segmentation, the performance demonstrates the superiority of the proposed approach as compared to the state-of-the-art (SOTA) methods.  
\end{abstract}


\IEEEpeerreviewmaketitle
\section{Introduction}
\IEEEPARstart{I}{mage} co-saliency detection (CoSOD) and object co-segmentation (CoSEG) are two important topics in computer vision. They often serves as a preliminary step for various down-streaming computer vision tasks, e.g., co-localization~\cite{DBLP:conf/cvpr/TangJLF14,DBLP:conf/eccv/JerripothulaCY16}, person re-identification~\cite{DBLP:conf/ijcai/LiuZZJ20} and 3D reconstruction~\cite{DBLP:conf/cvpr/ZhangHYH17,DBLP:conf/cvpr/MustafaH17}. These two tasks are highly relevant but different. For an image group, to detect (segment) co-occurring objects accurately, both these two tasks need model the synergistic relationship among the common objects. Although the co-occurring objects share the same semantic category, their explicit category attributes are unknown in CoSOD or CoSEG task. That is to say, CoSOD or CoSEG methods are not supposed to model the consistency relations of common objects by using the supervision of specific category labels or other information like temporal relations, which is quite different from video sequences tasks~\cite{DBLP:journals/tip/CongLFPHH19,DBLP:journals/tcsv/JerripothulaCY19}. These unique features make CoSOD or CoSEG an emerging and challenging task which has been rapidly growing in recent few years~\cite{9358006,DBLP:journals/tcsv/CongLFCLH19,DBLP:journals/tist/ZhangFHBL18,DBLP:journals/tip/ZhangLLWY21,DBLP:journals/mva/LiuD21,Zhang_2021_ICCV}. Different from CoSEG, CoSOD captures the saliency of the potential co-salient objects in the individual image (Intra-saliency). Consequently, our proposed method first considers the requirements to achieve high-quality co-saliency detection. 

{To detect co-saliency accurately, the core problem is how to stably model inter-saliency relations between an image group. Our previous~\cite{DBLP:conf/ijcai/0061STSS19} RCAN proposes a recurrent neural network based (RNN-based) model to address this problem. Compared to these 
CNN-based methods~\cite{DBLP:conf/ijcai/WeiZBLW17,DBLP:conf/cvpr/ZhangLL019,DBLP:conf/aaai/WangZLX19,DBLP:conf/nips/ZhangCHLZ20,9358006,GLNet} and graph-based methods~\cite{DBLP:conf/mm/0002JZT019,DBLP:conf/cvpr/ZhangLSLC020}, the RNN-based method is able to model more robust inter-saliency relations. Specifically, the main drawback of these CNN-based and graph-based methods is that they require constant input data, suffering from sub-group instability. When dealing with image groups containing a variable number of images, these CNN-based methods and Graph-based methods detect co-salient objects by dividing the image group into image pairs or image sub-groups. Since there is no principle way of dividing image groups, this strategy inevitably makes the overall training as well as testing process unstable, which influences the application of the co-salient object detection. On the contrary, the RNN-based method can adjust an unfixed number of images in each image group, and make use of all available information in an image group.} 

{After the work RCAN, some methods design sophisticated modules~\cite{DBLP:conf/nips/Jin0CZG20,DBLP:conf/eccv/ZhangJXC20} to address sub-group instability.} However, these sophisticated modules capture a single image attribute or pair of image-pair inter-saliency relations first, and then generate the final group features by directly adding these single attributes or pair of image-pair inter-saliency relations. Since there is much noise information in single image or image-pair features and the appearance as well as the location of co-salient object varies across different images or image-pair, only simple adding operation cannot capture these variations. While the RCAN proposes a novel recurrent co-attention mechanism to address this problem. 

\begin{figure}[]
\centering
\includegraphics[scale=0.6,width=8.5cm]{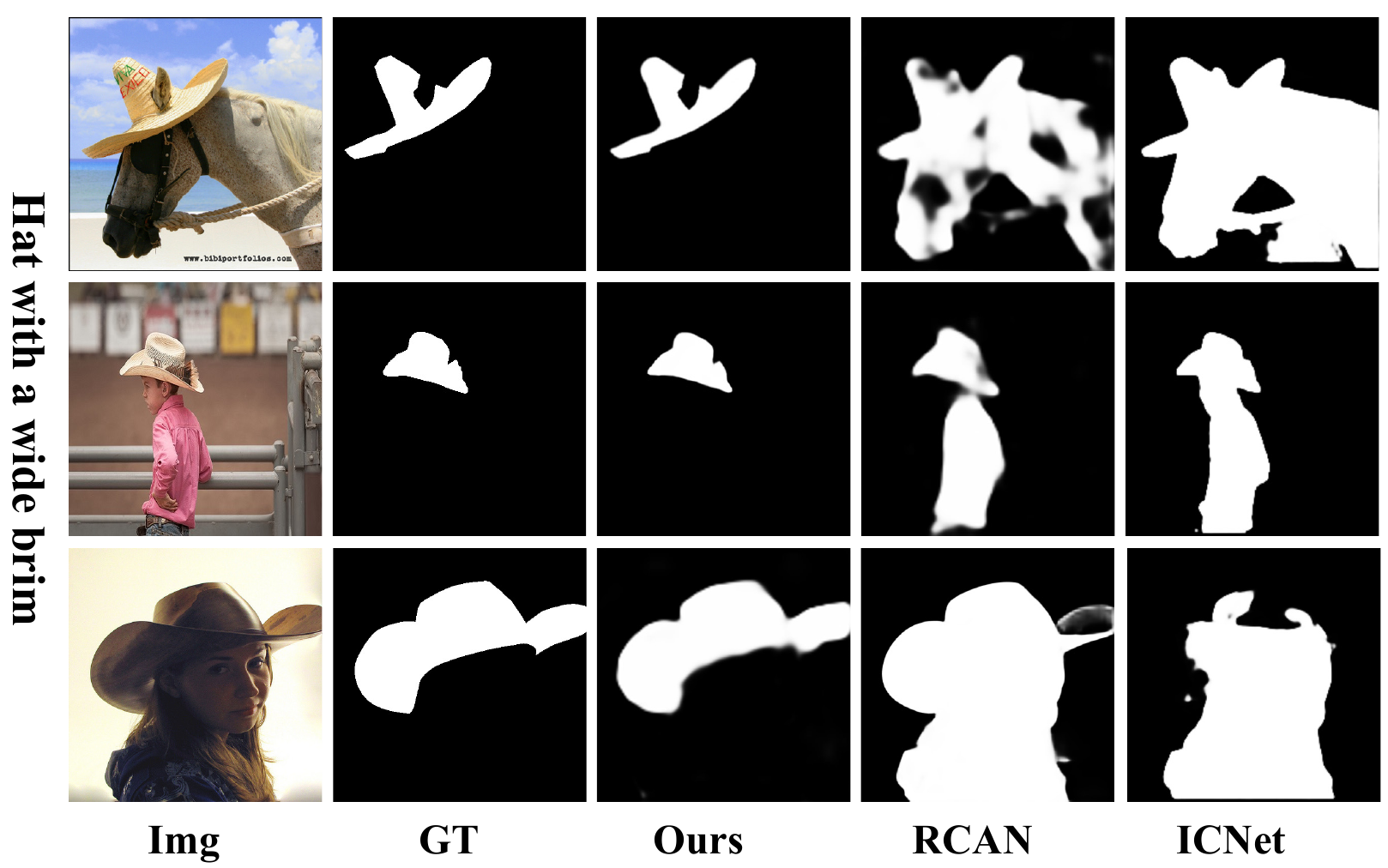}
\caption{ { Comparison with state-of-the-art methods RCAN~\cite{DBLP:conf/ijcai/0061STSS19} and ICNet~\cite{jin2020icnet}.} }
\label{introd}
\vspace{-0.5cm}
\end{figure}

In this paper, we revisit the sequential modeling for CoSOD task, then state the main problem of these RNN-based methods: order-sensitivity, which heavily affects the stable of proposed CoSOD model. Order-sensitivity is the inherent drawback of recurrent architecture, so RNN is only widely used in tasks which has strict order relations, such as natural language processing~\cite{wen2015semantically} (NLP) or video saliency detection~\cite{DBLP:journals/pr/ChenSZLL21}. In these tasks, the orders are pre-defined which are suitable for RNN. In contrast to NLP or video saliency detection, CoSOD tasks have no order relations between the images in an image-group. The RCAN cannot determine which orders are most suitable and different orders obviously affect its performance. 

To address the above problem, in this paper we first propose a multi-path stable recurrent unit (MSRU), containing dummy orders mechanisms (DOM) and recurrent unit (RU), which can handle the unstable drawback of RCAN. Similar to GRU~\cite{DBLP:conf/emnlp/ChoMGBBSB14}, RCAN contains a \textit{reset gate} and a \textit{update gate}, which help capture co-salient regions with help of spatial-channel-wise co-attention mechanism. As the appearance and the location of co-salient object varies large across images, and the previous attention mechanism in RCAN cannot capture long-range relations between different images, limiting the ability in capturing sufficient inter-saliency relations between different images. To address this problem, we design a novel non-local cross-attention (NLCA) in \textit{reset gate} and a novel co-attention feature projection module (CFPM) in \textit{update gate}, to fully mining common semantics in an image group. After capturing inter-saliency relations, we design another single image representation extraction branch (SIR) to process each image individually to learn intra-saliency of an image group. Finally, the outputs of these two branches are further fused through a non-local attention, which encourages rich interactions between the group and single representation to facilitate the robust co-saliency detection reasoning. 

To well supervise the network, in addition to the group-wise training objective proposed in RCAN, which is used to make full use of the interactive relationships of whole images in the training group, we design a cross-order contrastive loss (COCL) to further eliminate the effects from different input orders. In particular, our proposed COCL is capable of pulling close the group feature embeddings generated from different orders, which can further enhance the stability of our network. As can be seen in Figure.\ref{introd}, our proposed method can achieve best performance because of proposed MSRU and COCL.

Our main contributions can be summarized as follows:
\begin{itemize}

\item Taking the COSOD task as an example, we first investigate the capability of RNNs to model orderless sequence tasks. In addition, we state that order sensitivity is essential to network stability.
 
\item We propose a multi-path stable recurrent unit, which can collect multi-path group features generated from different orders for final stable group feature generation. In recurrent unit, we further design a novel non-local cross-attention (NLCA) and a novel co-attention feature projection module (CFPM) to fully mining common semantics in an image-group.

\item In additional to group-wise training objective, we design cross-order contrastive loss (COCL) to further eliminate the effects from different orders.  

\item We validate our model on five CoSOD datasets (CoCA, CoSOD3k, Cosal2015, iCoseg and MSRC), and three widely used datasets (Internet, iCoseg and PASCAL-VOC) for object co-segmentation, the performance demonstrates the superiority of the proposed approach as compared to the state-of-the-art (SOTA) methods.
\end{itemize}

{
The remainder of our paper is organized as follows. \textit{Section.\uppercase\expandafter{\romannumeral 2}} reviews the previous saliency detection, co-saliency detection, object co-segmentation methods and recurrent neural network. \textit{Section.\uppercase\expandafter{\romannumeral 3}} elaborates on the proposed network architecture. \textit{Section.\uppercase\expandafter{\romannumeral 4}} introduces the loss function in this work. \textit{Section.\uppercase\expandafter{\romannumeral 5}} provides extensive experimental results in comparison with SOTA methods and ablation studies of our proposed network. Finally, more discussion about \textit{the relations between co-saliency detection and object co-segmentation}, as well as \textit{more 
analysis of stability problem in the CoSOD/CoSEG task} are given in \textit{Section.\uppercase\expandafter{\romannumeral 6}}, respectively.

}

\section{Related Work}
We review relevant topics to the development of our approach, including saliency detection, co-saliency detection, object co-segmentation and recurrent neural networks.

\subsection{Saliency Detection}
Salient object detection (SOD) is a fundamental task in computer vision, which is derived with the goal of detecting and segmenting the most distinctive objects from visual scenes. Over the past decades, a large amount of SOD algorithms have been developed, which can be roughly classified into traditional methods and deep learning based methods. Traditional models \cite{DBLP:journals/pami/IttiKN98,DBLP:journals/pami/ChengMHTH15,DBLP:journals/ijcv/WangJYCHZ17,DBLP:conf/eccv/WangZLSQ16,DBLP:conf/iccv/KleinF11} detect salient objects by utilizing various heuristic
saliency priors with hand-crafted features. { Deep learning based models use various feature enhancement strategies to improve the ability of localization and awareness of salient objects~\cite{DBLP:conf/cvpr/WangSC019,DBLP:journals/pami/HouCHBTT19,DBLP:conf/cvpr/WuSH19,DBLP:conf/aaai/WangCZZ0G20,DBLP:conf/aaai/ChenXCH20,DBLP:journals/corr/abs-1911-11445,DBLP:conf/cvpr/PangZZL20,DBLP:conf/eccv/ZhaoPZLZ20,DBLP:conf/eccv/GaoTCLCY20,Tang_2020_ACCV,DBLP:conf/icassp/LiSTH19,DBLP:conf/icassp/TangLWXD21}, or take advantage of edge features to restore the structural details of salient objects~\cite{DBLP:conf/cvpr/WangZSHB19,DBLP:conf/cvpr/WangZSHB19,DBLP:conf/iccv/ZhaoLFCYC19,DBLP:conf/iccv/WuSH19,DBLP:conf/cvpr/ZhouXLCY20}. Different from the above methods, some methods~\cite{DBLP:conf/cvpr/WangZWL0RB18,DBLP:conf/cvpr/QinZHGDJ19,Tang_2021_ICCV} consider leveraging predict-refine architecture or the fixation prediction framework~\cite{DBLP:conf/cvpr/WangSDB18,DBLP:journals/pami/WangSDBY20} to generate fine salient objects. } Beyond of the scope of the paper, more detailed introduction of salient detection can be referenced in recent surveys~\cite{DBLP:journals/cvm/BorjiCHJL19,DBLP:journals/tcsv/CongLFCLH19}. 

\subsection{Co-saliency Detection}
Compared to SOD task, CoSOD needs to model inter-saliency relations among an image group. Therefore CoSOD task is more challenging than saliency detection. To model inter-saliency relations between an image group, conventional approaches~\cite{DBLP:journals/tip/LiN11, DBLP:conf/icip/Chen10a,DBLP:conf/mm/CaoCTF14, DBLP:conf/eccv/JerripothulaCY16, DBLP:conf/cvpr/ChangLL11,DBLP:journals/spl/LiFL015,DBLP:journals/tip/FuCT13,liu2014co} utilize handcrafted features, such as color, texture and SIFT descriptors etc., and these methods rely on researcher’s prior knowledge to model the interaction between the group images, like inter-image saliency. However, low-level features and fixed hand-designed interaction models are too subjective to face the multiple challenges including background clutter, appearance variance of co-salient object across images, and similarity between co-object and non-common object, etc. Recently deep-based models simultaneously explore the intra-saliency and inter-image consistency in a supervised manner with different approaches, such as concatenation operation~\cite{DBLP:conf/ijcai/WeiZBLW17,DBLP:conf/nips/ZhangCHLZ20}, graph convolution networks (GCN) \cite{DBLP:journals/tcsv/TangLKSD22,DBLP:conf/mm/0002JZT019, DBLP:conf/cvpr/ZhangLSLC020}, self-learning methods \cite{DBLP:journals/pami/ZhangMH17, DBLP:journals/tnn/ZhangHHS16}, co-clustering \cite{DBLP:journals/tip/YaoHZN17} or Transformer-based methods~\cite{DBLP:journals/corr/abs-2104-14729}. However, the main drawback of these methods is that they suffer from sub-group instability. While the size of each group is not fixed in real-world scenarios as well as the experimental co-saliency dataset, so only partial inter-saliency relationships will be captured, limiting the robustness of the model. To address this problem, some methods design sophisticated modules, like gradient feedback~\cite{DBLP:conf/eccv/ZhangJXC20}, correlation techniques~\cite{DBLP:conf/nips/Jin0CZG20}, to adjust unfixed number of images in each image group. In general, these two methods simply use the adding (fusion) to generate final inter-saliency relations from sub inter-saliency relations. While the location of co-salient region and noisy region vary in differnet image, only simply adding operation cannot make fully interaction between differnet images and retain co-salient regions. RCAN~\cite{DBLP:conf/ijcai/0061STSS19} proposes RCAU mechanism, which can better suppress non-salient background and retain co-salient regions. As we know, recurrent architecture  makes the network sensitive to input orders.
In this paper, we propose MSRU to make our proposed network stable and be less sensitive to input orders. For more about CoSOD tasks, please refer to~\cite{9358006,DBLP:journals/tcsv/CongLFCLH19,DBLP:journals/tist/ZhangFHBL18}.

\subsection{Object Co-Segmentation}
The concept of co-segmentation was first introduced by the Rother~\cite{DBLP:conf/cvpr/RotherMBK06}, who used histogram matching to simultaneously segment out the common object from a pair of images. Following this work, a number of researchers have made further efforts to develop more effective object co-segmentation models by comparing foreground color histograms~\cite{DBLP:conf/eccv/VicenteKR10} or adopting more diverse features like Gabor filters~\cite{DBLP:conf/iccv/HochbaumS09} and SIFT~\cite{DBLP:conf/cvpr/RubinsteinJKL13}. In order to better explore the correspondence relationship among common objects, some existing methods \cite{DBLP:conf/iccv/DaiWZZ13,DBLP:journals/ijcv/SunP16,7156153,DBLP:journals/tmm/JerripothulaCY16,DBLP:conf/cvpr/VicenteRK11,DBLP:conf/mmm/HuSLYL17} additionally introduced prior constraints to better distinguish them from the undesired image backgrounds. However, these methods cannot obtain robust performance in real-world scenarios, where the handcrafted low-level features are too subjective to face the multiple challenges including intra-class variations and background clutters and the predefined prior knowledge cannot always provide adequate and precise constraint on the common objects. 

Recent researches~\cite{DBLP:conf/ijcai/HsuLC18,DBLP:conf/cvpr/QuanHZN16,DBLP:conf/ijcai/YuanLW17} use deep visual features to improve object co-segmentation and they also try to learn more robust synergetic properties among images in a data driven manner. Yuan~\cite{DBLP:conf/ijcai/YuanLW17} introduced a DNN-based dense conditional random field framework for object co-segmentation by cooperating co-occurrence maps which are generated using selective search~\cite{DBLP:journals/ijcv/UijlingsSGS13}. Hsu~\cite{DBLP:conf/ijcai/HsuLC18} proposed a DNN-based method which uses the similarity between images in deep features and an additional object proposals algorithm~\cite{DBLP:conf/eccv/KrahenbuhlK14} to segment the common objects. These  methods achieved state-of-the-art results by substituting the features learned by DNN for engineered features.  However, as feature learning and object segmentation are somehow separated in these approaches, the learned features are not tailored for segmenting the co-occurring objects, resulting in suboptimal performance. The very recent works~\cite{DBLP:journals/corr/abs-1810-06859,DBLP:journals/corr/abs-1804-06423} proposed end-to-end deep learning methods for co-segmentation by integrating the process of feature learning and co-segmentation inferring as an organic whole. By introducing the correlation layer~\cite{DBLP:journals/corr/abs-1804-06423} or a semantic attention learner~\cite{DBLP:journals/corr/abs-1810-06859}, they can utilize the relationship between the image pair and then segment the co-object in a pairwise manner.  However, their siamese network structures limit their use of group-wise information which contains more sufficient synergetic relationships than image pairs. Consequently, co-segmenting common objects from image pairs has very limited robustness and practical application value when extending beyond pairwise relations. Unlike the previous methods, by introducing the recurrent architecture, our co-segmentation network is able to make use of all available information including individual image properties and the group-level synergetic relationships to meet the need for real-world applications. Recently, method~\cite{DBLP:journals/tip/ZhangLLWY21} introduces region correspondence module which can help the network handle unfix input, while the drawback of this work is it omits how to make the model robust to different orders. One closely related topic to object co-segmentation is co-saliency detection, which aims at generating co-saliency maps for each of the images from the given image collection to highlight the common and salient objects. Compared with co-saliency detection, object co-segmentation only aims at segmenting the co-occurring objects without constraining those objects to be (co-)salient. By altering the datasets, we are able to use one network to address both CoSOD and CoSEG problems simultaneously in this paper. {By the way, we find that the existing method~\cite{DBLP:journals/tip/TsaiLHQL19} also tries to design a unified network to address these two problems. More details can be seen in Section.\ref{experimental} and Section.\ref{discussion}.}

\subsection{Recurrent Neural Network}
RNN have been widely used in NLP(e.g., \cite{wen2015semantically}) and speech recognition (e.g., \cite{DBLP:conf/fpga/HanKMHLLXLYWYD17}) to understand sequence data. The most popular variants of RNN included Long Short-Term Memory (LSTM~\cite{DBLP:journals/neco/HochreiterS97}) and Gated Recurrent Unit (GRU \cite{DBLP:conf/emnlp/ChoMGBBSB14} ). The common LSTM unit is composed of a cell and three gates (forget gate, input gate and output gate), which is designed to be capable of learning long term dependencies. Recently, the RNN (especially the LSTM and GRU) has been introduced in spatiotemporal tasks (known as convolutional RNN) for precipitation nowcasting \cite{DBLP:conf/nips/ShiCWYWW15}, pattern recognition \cite{DBLP:journals/corr/abs-1904-10709,DBLP:journals/ijon/LvXC19}, trajectory prediction~\cite{DBLP:conf/itsc/AltcheF17}, medical image analysis\cite{DBLP:conf/miccai/YangXXZXCPGTCMC17,DBLP:journals/corr/abs-1709-02081,xu2019deep} and video saliency detection~\cite{DBLP:journals/corr/abs-1709-06316,DBLP:conf/cvpr/WangSGCB18,DBLP:journals/pr/ChenSZLL21}, et cetera. However, all these works process sequential data, so the variation between different images is little. While in co-saliency detection task, the location of co-salient objects varies greatly in different images. And the different gates in LSTM or GRU cannot handle these complex variations. To address this problem, RCAN propose co-attention mechanism to well model these inter-relations. However, the main drawback of the RCAN is that it makes the network sensitive to input orders. Hence, in this paper, we further design a MSRU to address order-sensitive problem. 

\begin{figure}[!t]
\centering
\includegraphics[scale=0.6]{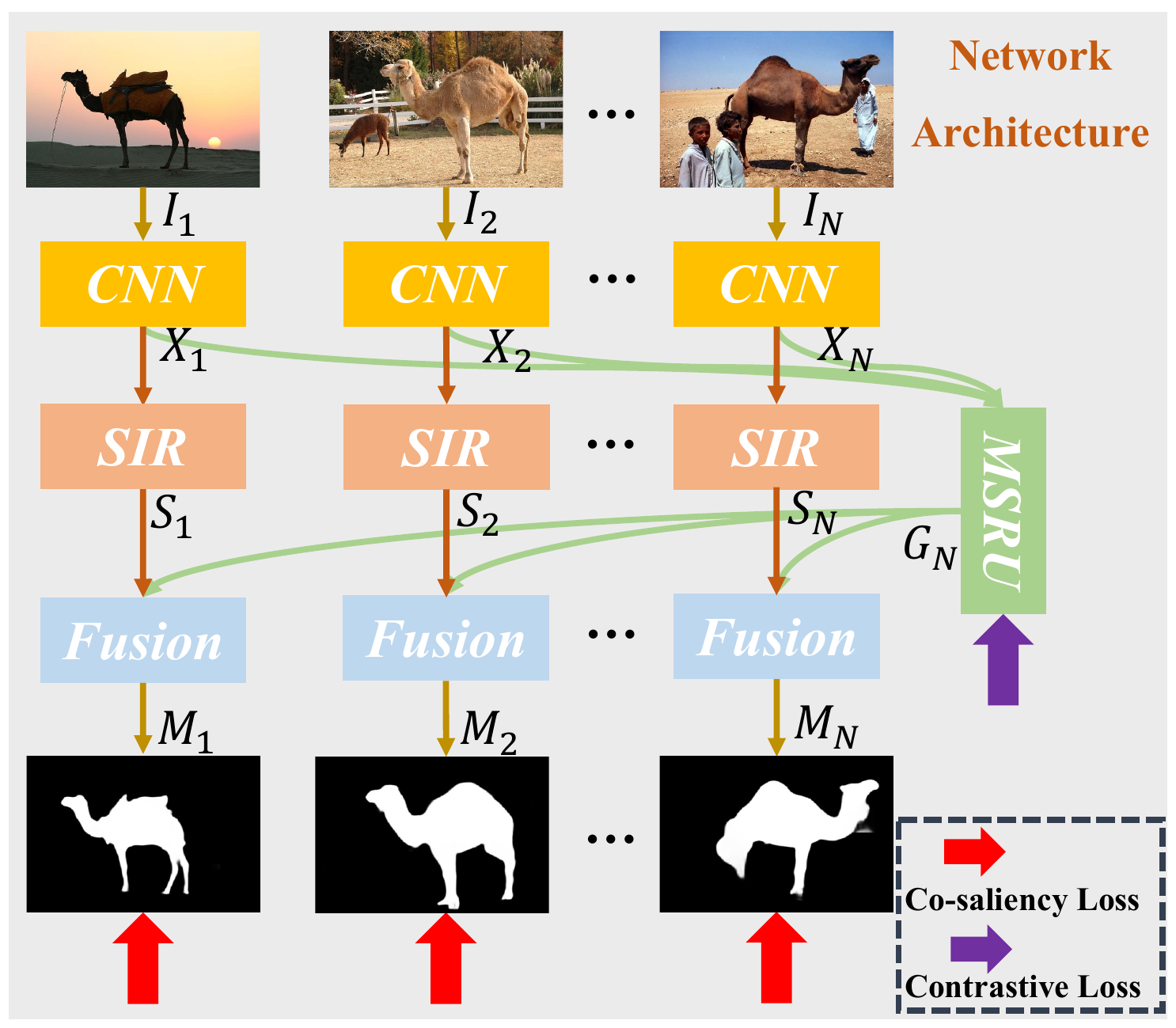}
\caption{Illustration of the proposed recurrent network architecture for co-saliency detection and object co-segmentation.}
\vspace{-0.25cm}
\label{Architecture}
\end{figure}

\section{Proposed Method}

\subsection{Network Overview}
The overall architecture of the proposed approach is illustrated in Figure.\ref{Architecture}. Co-saliency detection aims at discovering the co-occurring object masks $\mathcal{M} = \{M_{n}\}^N_{n=1}$ from a group of $N$ relevant images $\mathcal{I} = \{I_{n}\}^N_{n=1}$. For an input image group with an arbitrary size, our network first uses CNN to extract the semantic features of all images. Then the single image representation (SIR) branch processes each image individually to learn the intra-saliency $\{S_{n}\}^N_{n=1}$. Meanwhile, the multi-path stable recurrent unit (MSRU) recurrently explores all images in the image group to learn the robust group representation $G_N$. Finally, the outputs of these two branches are further fused through a non-local fusion module for robust co-saliency detection. The loss function of our proposed network contains co-saliency loss and cross-order contrastive loss (COCL).  

\subsection{Intra-saliency Learning}
As a basic rule in co-saliency detection, it is important to learn the unique properties of each image to capture potential co-occurring objects in the individual image. For each image $I_{n}$ in the input group $\mathcal{I}$, we first use a pre-trained VGG16~\cite{DBLP:journals/corr/SimonyanZ14a} to extract semantic features. Following~\cite{DBLP:journals/pami/HouCHBTT19,DBLP:conf/iccv/ZhaoLFCYC19}, we connect another side path to the last pooling layer in VGG-16. Hence, we obtain six side features Conv1-2, Conv2-2, Conv3-3, Conv4-3, Conv5-3 and Conv6-3 from the backbone network. {  Before sending the side features extracted from VGG to the SIR, we first use a $1\times1$ convolutional operation to reduce their channel numbers for saving computation. In this paper, we set channel number $C=64$ suggested by works~\cite{DBLP:conf/iccv/WuSH19,DBLP:conf/iccv/TangLZDS21}. } For simplicity, we name side feature Conv6-3 as ${X_{n}}\in \mathcal{R}^{ C \times H \times W}$. So the side features of image group can be written as: $\mathcal{X} = \{X_{n}\}^N_{n=1}$. Then we construct the SIR block on $\mathcal{X} = \{X_{n}\}^N_{n=1}$, to capture the intra-saliency $\mathcal{S}=\{S_{n}\}^N_{n=1}$ for each image. {  The SIR contains three convolutional blocks, and each block contains a $3 \times 3$ convolutional operation with $stride=1$, followed by a batch normalization and a ReLU activation.} Note that we capture intra-saliency and inter-saliency relations on three levels of backbone network (Conv4-3, Conv5-3 and Conv6-3). In this paper, we only show the intra-saliency learning and inter-saliency relations capturing in Conv6-3 for simplicity.

\begin{figure}
\centering
\includegraphics[scale=0.46]{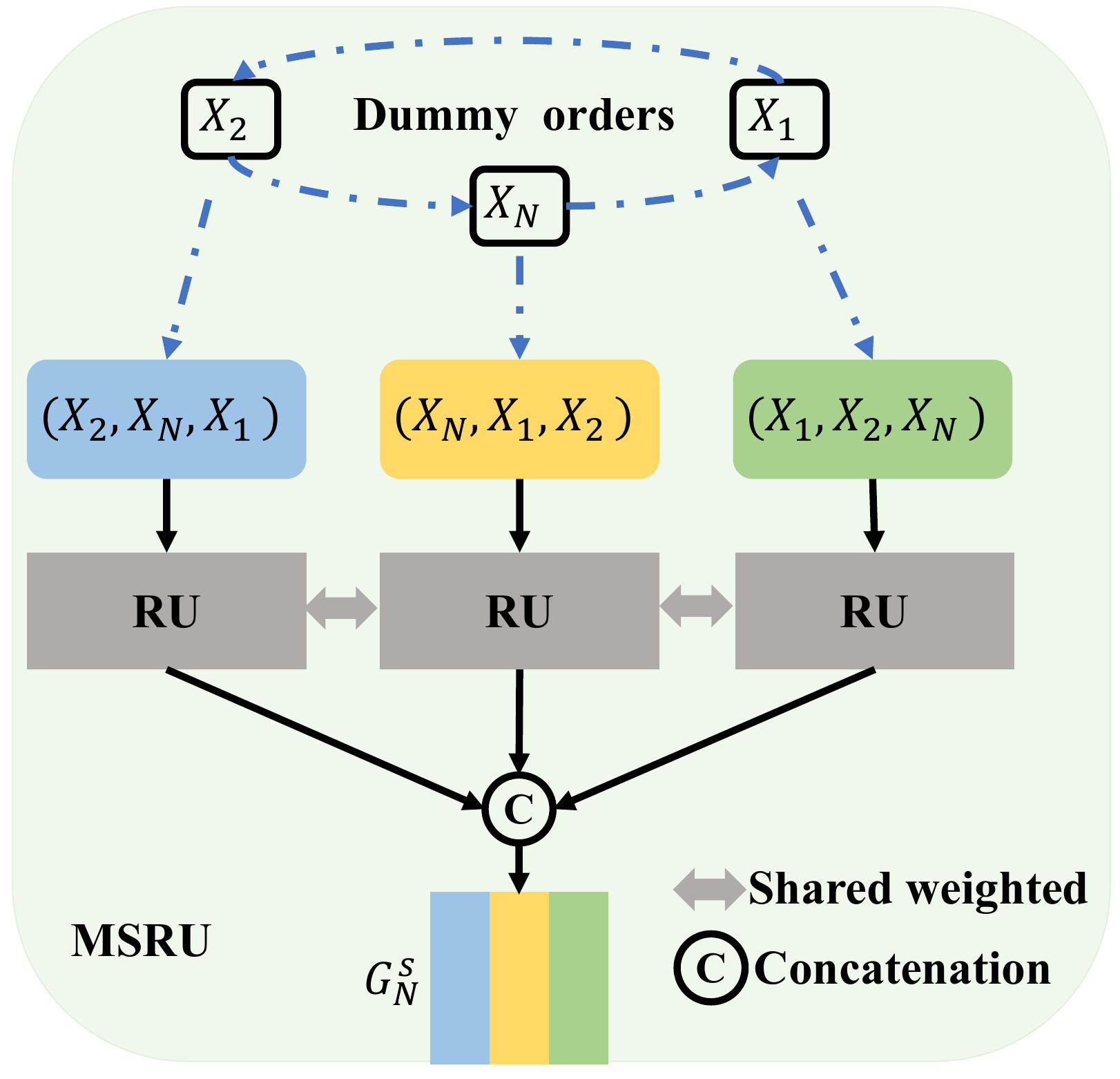}
\caption{The proposed multi-path stable recurrent unit (MSRU), which contains contains dummy orders mechanism (DOM) and a recurrent unit (RU). }
\label{DOM}
\vspace{-0.25cm}
\end{figure}

\subsection{Inter-saliency Relations Capturing}
As images within an image group are contextually associated with each other in different ways such as common objects, similar categories, and related scenes, learning a robust group representation which contains the relevance and interaction between group images is extremely important for co-saliency referring. Our previous work RCAN~\cite{DBLP:conf/ijcai/0061STSS19} proposes to use a RCAU to learn the group representation for an arbitrary size group $\mathcal{I}$. Since the main drawback of the previous RCAU is that it makes the network sensitive to input orders, leading to unstable training and inferring procedures. To address this problem, we propose a new multi-path stable recurrent unit (MSRU, Figure.\ref{DOM}) which can collect features from different orders for final group features generation. Next, we will describe the MSRU in detail. 

The feature representations of an image sequence that contains $N$ images are written as:
\begin{equation}
\{ X_1, X_2, ..., X_N\}.
\end{equation}
Previous RCAU gradually update each single image feature representation to the final group feature, and the performance is easily influenced by the order of input images. An intuitive idea to solve this problem is to generate all the different orders of an image group, and then concatenate group features from these different orders for final group representations. However, this idea is impractical because totally $N!$ orders are generated. Therefore, we propose a compromise approach in this paper. Specifically, the original sequence is first divided into several sub-groups by slide window with $stride = 1$: 
\begin{equation}
\{ \{X_{1},X_{2},X_{3}\},...,\{X_{N-1},X_{N},X_{1}\},\{X_{N},X_{1},X_{2}\} \}.
\label{eq2}
\end{equation}
Here the window size is set as 3, and we treat the image sequence as a cycle. Hence, the group features of each sub-group is denoted as $\{G_n^s\}_{n=1}^N$, which can be written as:
\begin{equation}
\{ G_1^s,...,G_{N-1}^s,G_N^s\}.
\label{eq3}
\end{equation}
In this paper, we first generate multi-path feature representations of each sub-group $G_n^s$ from three different orders, then use $\{G_n^s\}_{n=1}^{N}$ to generate the final group feature $G_N$ (Figure.\ref{Architecture}).
\begin{figure}
\centering
\includegraphics[scale=0.48]{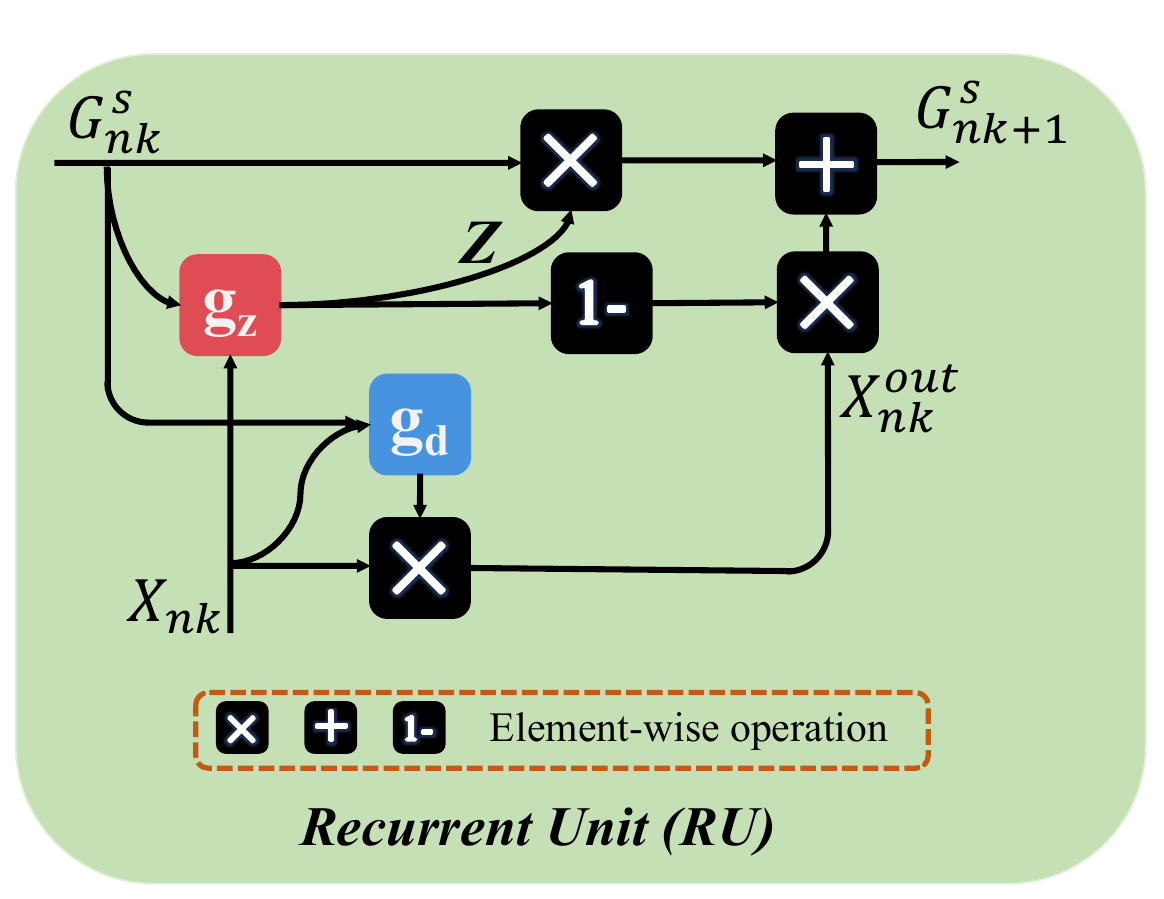}
\caption{The architecture of the proposed recurrent unit (RU).}
\label{RU}
\end{figure}
To generate the multi-path group feature representations of each sub-group $G_n^s$, we first propose a MSRU, which contains dummy orders mechanism (DOM) and a recurrent unit (RU). As can be seen in Figure.\ref{DOM}, if a sub-group $G_N^s$ only contains three images $ \{X_{N},X_{1},X_{2}\}$, the DOM would generate three different orders \{ $``X_{1} \to X_{2} \to X_{N}"$, $``X_{2} \to X_{N} \to X_{1}"$, $``X_{N} \to X_{1} \to X_{2}"$\} of this group. Then we propose a RU to generate three group feature representations of these three different orders. Finally, we concat these three group feature representations to achieve multi-path group feature representations of this sub-group $G_N^s$.  

Before introducing our proposed RU, we make some modifications on Eq.\ref{eq2} and Eq.\ref{eq3}, which can help readers better understand the process of RU. In Eq.\ref{eq2}, we can get several
sub-groups, and rewrite their notations as:
\begin{equation}
\{ \{X_{(1)1},X_{(1)2},X_{(1)3}\},...,\{X_{(N)1},X_{(N)2},X_{(N)3}\} \}.
\end{equation}
The images in each sub-group can be written as $X_{nk}$. $X_{nk}$ means the feature of $k$-th image in $n$-th sub-group, where $k\in[1,3]$ and $n\in[1,N]$ in this paper. So the group feature of one order belong to each sub-group can be written as $G^{s}_{nK}$ ($K=3$). Although $G^{s}_{nK}$ is the group feature of one order, to avoid the abuse of notations, $G^{s}_{nK}$ also means the 
concated multi-path group feature representation of sub-group $G_n^s$. Next we will show the details of our proposed RU. 

The two key modules of previous RCAU are \textit{reset gate} and \textit{update gate}. The goal of \textit{reset gate} ($g_d$) is to use the synergetic relationships between the group feature and the current image to suppress the noise data in current image. The previous RCAU only uses a convolutional operation to achieve this purpose. However, this simple convolutional operation cannot capture long-range relations between group feature and current image feature. So if the co-salient regions of these two features vary largely, simple convolutional operation cannot well suppress the noise data. Inspired by non-local network, we design a new non-local cross-attention (NLCA) to fully suppress the noise data. Specifically, the input of NLCA is the group feature representation $G_{nk}^s \in \mathbb{R}^{ C \times H \times W }$ and single image representation $X_{nk} \in \mathbb{R}^{ C \times H \times W }$. In the first step, $G_{nk}^s$ is initialized with $X_{nk}$. For \textit{Query} branch, we first add a $1 \times 1$ convolution layer on $X_{nk}$ and reshape the feature to $\mathcal{R}^{C\times L}$, where $L = H \times W$. Meanwhile, for \textit{Key} branch, we also use a $1 \times 1$ convolution layer on $G_{nk}^{s}$ and reshape the feature to $\mathcal{R}^{C\times L}$. After that, we perform a matrix multiplication between the transpose of $X_{nk}$ and $G_{nk}^{s}$, then apply a \textit{softmax} function to calculate the spatial attention map $\textbf{A} \in \mathcal{R}^{N \times N}$. Each pixel value in $\textbf{A}$ is defined as:
 \begin{equation}
A(i,j) = \frac{ exp( (X_{nk})^i \cdot (G_{nk}^{s})^j ) }{\sum_{j=1}^{L}exp( (X_{nk})^i \cdot (G_{nk}^{s})^j ) },
\end{equation}
where $i \in [1,L]$, $A(i,j)$ measures the $j^{th}$ position in the group feature impact on $i^{th}$ position in single image feature. Meanwhile, like \textit{Key} branch, we generate feature $\hat{G}_{nk}^{s}$ from \textit{Value} branch and perform a matrix multiplication between $\textbf{A}$ and the transpose of $\hat{G}_{nk}^{s}$ to get denoised feature $X_{nk}^{out} \in \mathcal{R}^{C \times L}$, which is defined as:
 \begin{equation}
X_{nk}^{out}(i) = \sum_{j=1}^{N}A(i,j)\hat{G}_{nk}^{s}(j),
\end{equation}
Finally, we reshape $X_{nk}^{out}$ to $ \mathcal{R}^{ C \times H \times W }$, and add it with $X_{nk}$.

\begin{figure}
\centering
\includegraphics[scale=0.48]{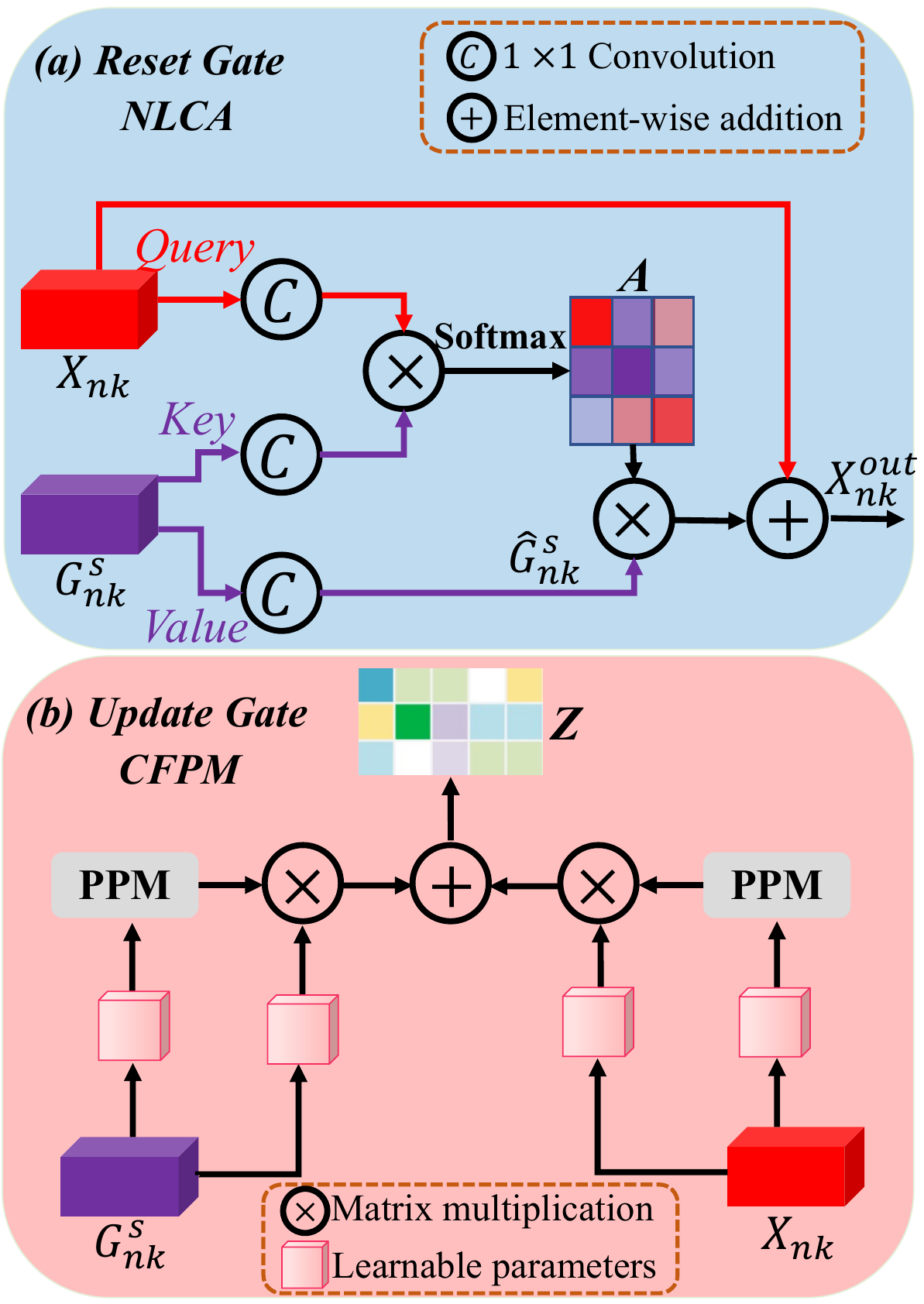}
\caption{(a) is the architecture of proposed \textit{reset gate} $g_d$. (b) is the architecture of proposed \textit{update gate} $g_z$. }
\label{Gate}
\end{figure}

The goal of \textit{update gate} ($g_z$) is to determine what group information should be retained in group feature and what new information should be updated from current image feature. Previous RCAU uses co-attention mechanism to explore the spatial-channel-wise variation of the co-salient object between group feature $G_{nk}^s$ and image feature $X_{nk}$. However, the appearance as well as the location of co-salient object varies across different image and group features. Therefore, it is difficult to capture the consistency information between two differently distributed features with a spatial-channel-wise attention map. So we propose a co-attention feature projection module (CFPM) to project image and group features to the common feature subspace, which can help bridge the gap between group feature and image feature. At first, we employ pyramid pooling module (PPM~\cite{DBLP:conf/iccv/ZhuXBHB19}) to reduce the dimension of feature maps $G_{nk}^s$ and $X_{nk}$ to save the computational cost. The PPM is composed of four-scale feature bins, which are then flattened and concatenated to form a matrix of size $C_1 \times L_g$, $L_g \ll HW$, $C_1 = C / 2$. Here, the sizes of the feature bins are set to $1 \times 1$, $3 \times 3$, $6 \times 6$ and $8\times 8$, respectively. Thus, the self-affinity matrixes of $G_{nk}^s$ and $X_{nk}$ can be calculated as:

\begin{equation}
\begin{aligned}
& A_G = (PPM(W_G^1G_{nk}^s))^\top(W_G^2G_{nk}^s),\\
& A_S = (PPM(W_S^1X_{nk}))^\top(W_S^2X_{nk}),
\end{aligned}
\end{equation}
where $A_G$ and $A_S$ denote the feature-specific similarity matrixes. Their sizes are fixed to $L_g \times (HW)$ through the PPM, which is asymmetric. $W_G^1$, $W_G^2$, $W_S^1$ and $ W_S^2 \in \mathbb{R}^{C_1 \times C}$,  indicate the learnable parameters. We further combine these two matrices as follows:
\begin{equation}
\begin{aligned}
Z = softmax((A_G+A_S)^\top).
\end{aligned}
\end{equation}

Finally, the row-wise normalized matrix $Z$ $\in$ $\mathbb{R}^{ L_g \times (HW)}$ is used to assist the update of group and image features:
\begin{equation}
\begin{aligned}
& \widetilde{G}_{nk}^s = Z(W_G^3 G_{nk}^s)^\top, \\
& \widetilde{X}_{nk} = (1-Z) (W_S^3 X_{nk}^{out} )^\top.
\end{aligned}
\end{equation}
We add $\widetilde{G}_{nk}$ and $\widetilde{X}_{nk}$ to get the features $G_{nk+1}^s \in  \mathbb{R}^{C \times H \times W}$:
\begin{equation}
G_{nk+1}^s = \widetilde{G}_{nk}^s + \widetilde{X}_{nk}.
\end{equation}
In this paper, the above process would be repeated three times to get group features of one order. Finally, we concat the group features from three different orders to generate multi-path group feature ($G^{s}_{nK}$ or $G_n^s$) of each sub-group.   

After generating sub-group features $ \{G_n^s\}_{n=1}^N$ from $\mathcal{X} = \{X_{n}\}^N_{n=1}$, we apply another RU to generate the final group feature $G_N$ from $\{G_n^s\}_{n=1}^N$, which is written as:
\begin{equation}
G_N = RU( \{G_n^s\}_{n=1}^N ).
\end{equation}

\subsection{Co-saliency Detection with Fused Representation}
As described previously, the group feature is then broadcasted to each image, which allows the network to leverage the synergetic information and unique properties between the images. So the interaction of group representation $G_N$ and single representation  $\{S_{n}\}^N_{n=1}$ are sufficiently exploited to facilitate the robust co-saliency reasoning. Thus, inspired by classic non-local network~\cite{DBLP:conf/cvpr/0004GGH18}, we propose a non-local fusion module to well fuse single representations and group features, and get the final co-saliency maps $\mathcal{M} = \{M_{n}\}^N_{n=1}$.

\section{Loss Function}
To well optimize the network, we propose the co-saliency loss $L_{co}$ and cross-order contrastive loss (COCL) $L_{cocl}$. Following the previous work RCAN, $L_{co}$ contains cross-entropy loss and perceptual group-wise loss, which can help the network in achieving a good co-saliency result. Moreover, we design a COCL $L_{cocl}$ that can further improve order-sensitive problem by pulling close the feature embedding generated from different input orders, resulting in a more stable inference and training process. 

\begin{figure}[]
\centering
\includegraphics[scale=0.4]{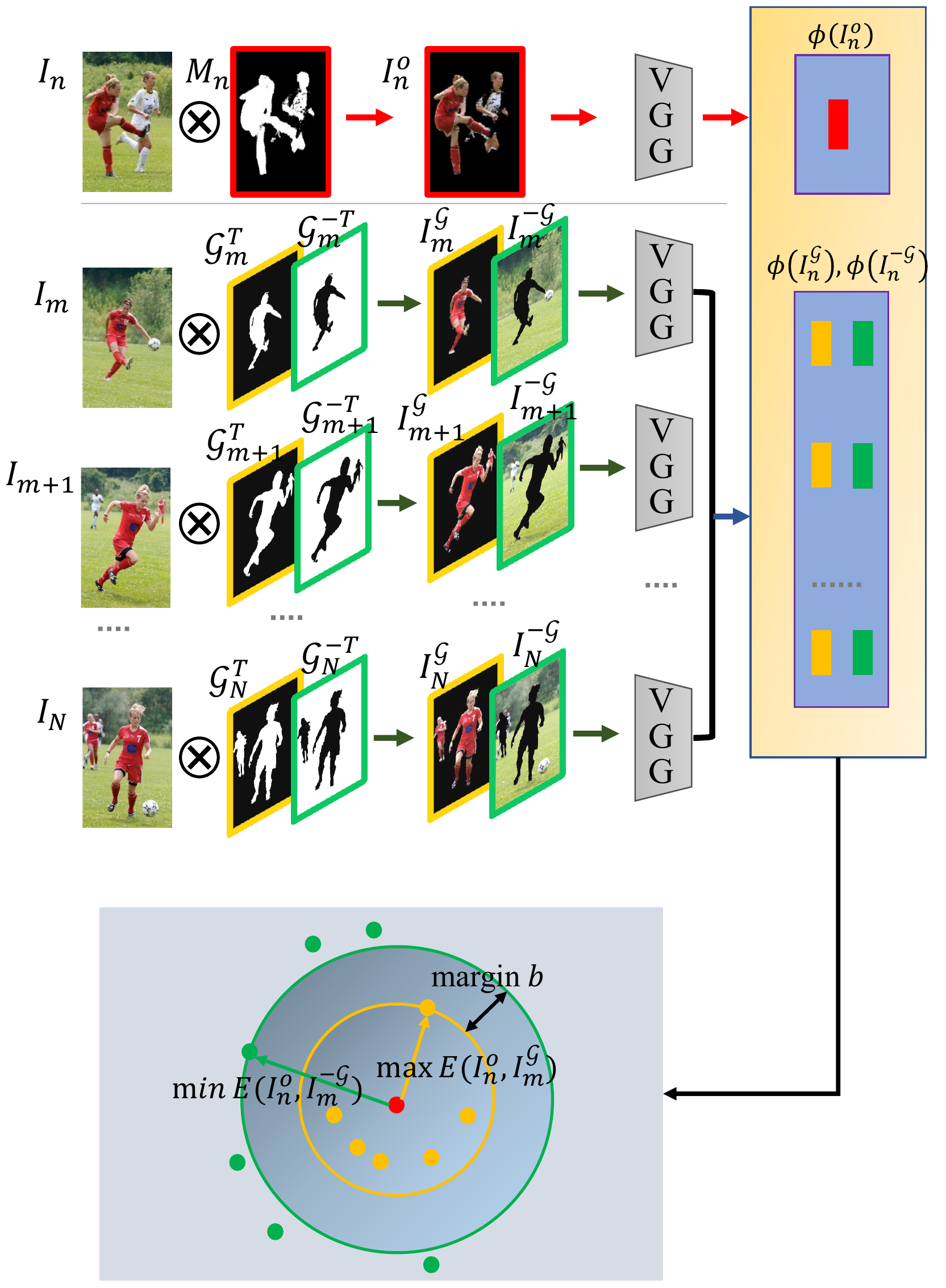}
\caption{Illustration of the group-wise training objective.}
\label{Co-Loss}
\end{figure}

\subsection{Co-saliency Loss}
Let $\mathcal{I}=\{I_{n}\}^N_{n=1}$ and their groundtruth  $\{\mathcal{G}^{T}_{n}\}^N_{n=1}$ denote a collection of training samples where $N$ is the number of images. After co-saliency detection, the co-saliency results are $\{M_{n}\}^N_{n=1}$. We use the cross-entropy loss as the individual supervision for each image $I_{n}$: 
\begin{equation}
\setlength{\abovedisplayskip}{4pt}
L_{s} = -\big(\ \mathcal{G}^{T}_{n}log(M_{n}) + (1-\mathcal{G}^{T}_{n})log(1-M_{n}) \ \big).
\end{equation}
In addition to the cross-entropy losses, we propose a perceptual group-wise training objective to further explore the interactive relationships of whole images in the training group. Two criteria are jointly considered in the design of group-wise training objective, including 1) high cross-image similarity between the co-occurring objects and 2) high distinctness between the detected co-occurring objects and the rest of the images like background and non-common objects. We apply triplet loss as the group-wise constraint. Specifically, for a image $I_n$, we can generate three masked images with its co-salient mask $M_n$ and groundtruth $\mathcal{G}^T_n$
\begin{equation}
\setlength{\abovedisplayskip}{4pt}
I^o_n = M_n \otimes I_n, ~ I^{\mathcal{G}}_n = \mathcal{G}^T_n \otimes I_n ~ and ~ I^{-\mathcal{G}}_n = (\mathcal{G}^{-T}_n)\otimes I_n,
\end{equation}
where $\otimes$ denotes element-wise multiplication and $\mathcal{G}^{-T}_n = 1-\mathcal{G}^T_n$. The masked image $I^o_n$ means our current detected co-salient objects of $I_n$ while image $I^{\mathcal{G}}_n$ and $I^{-\mathcal{G}}_{n}$ mean the real co-salient objects and non-common regions of $I_n$. Then we apply the perceptual extractor~\cite{DBLP:conf/cvpr/GatysEB16} $\phi$ to all masked images $\{I^o_n,I^{\mathcal{G}}_n,I_{n}^{-\mathcal{G}}\}^N_{n=1}$ and obtain their corresponding perceptual features $\{ \phi(I^o_n), \phi(I^{\mathcal{G}}_n), \phi(I^{-\mathcal{G}}_n)\}^N_{n=1}$. We apply triplet loss $L_c$ on each $I^o_n$ as the group-wise training objective as shown in Figure 4, formulated as:
\begin{equation}
\begin{split}
L_c = \frac{1}{N-1}\sum_{m \neq n}\big[b &+ max \ \  E(I^o_n, I^{\mathcal{G}}_m) \\ &-  min \ \ E(I^o_n,I^{-\mathcal{G}}_m)\big]_{+},
\end{split}
\end{equation}
where $b$ is the margin and $E(\cdot,\cdot)$ denotes the Euclidean distance between two feature vectors. The group-wise training objective uses the hinge function $[b+\bullet]_{+}$ to force co-saliency result $M_n$ to be more similar to real co-saliency objects than non-common regions. In co-saliency task, it can be beneficial to pull together co-occurring objects as much as possible. For this purpose, it is possible to replace the hinge function by a smooth approximation using the softplus function: In$\big(1+exp(\bullet)\big)$. The softplus function has similar behavior to the hinge, but it decays exponentially instead of having a hard cut-off, we hence refer to it as the soft-margin formulation. So the total co-saliency loss can be written as:
\begin{equation}
L_{co} = L_s + L_c. 
\end{equation}

\begin{figure}[]
\centering
\includegraphics[scale=0.5,width=7.0cm]{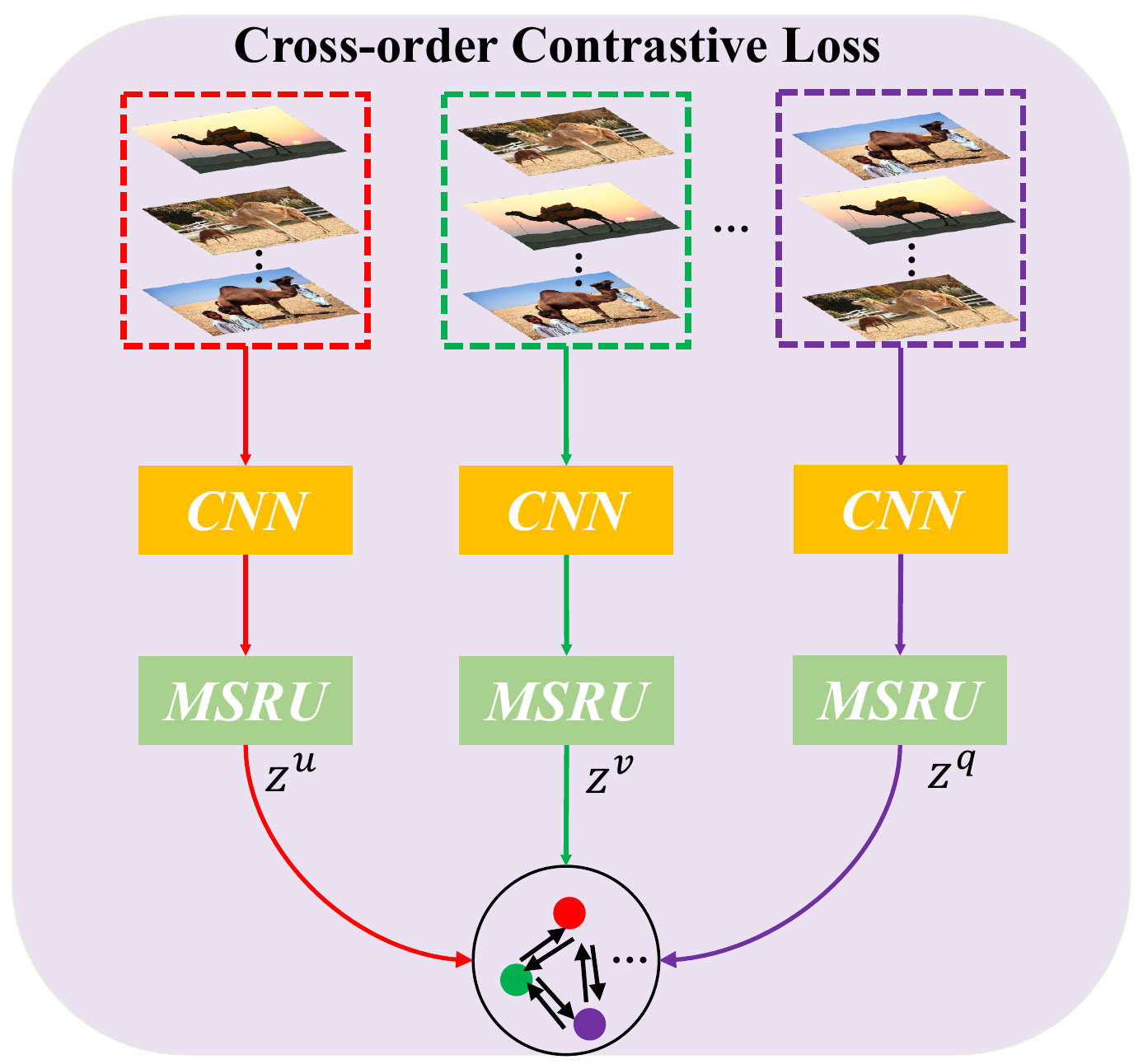}
\caption{Illustration of the contrastive loss. We set $q=10$ in this paper.}
\label{Contrastive}
\end{figure}

\subsection{Cross-order Contrastive Loss}
In this paper, we first propose MSRU to make our network insensitive to the input orders. A stable CoSOD network means different input orders can generate similar group features. This means we should pull close embedding spaces of different group features. So we design a cross-order contrastive loss (COCL) to achieve this purpose. However, existing contrastive losses (e.g. \cite{DBLP:conf/cvpr/He0WXG20, DBLP:conf/icml/ChenK0H20,DBLP:conf/nips/GrillSATRBDPGAP20}) cannot be directly applied to CoSOD task, because their aim is to help the network distinguish between feature embeddings of different inputs. Data augmentations are used to embed one-view data into different spaces, which helps the network learn the distinctive feature embedding of one-view data. While in CoSOD task, we leverage contrastive loss to further pull close different group features and make our proposed model insensitive to the input orders. So we propose the cross-orders consistent mining, leveraging the high similarity of samples in one order to guide the learning process in another order. It excavates positive pairs across different orders according to the embedding similarity to promote knowledge exchange among different orders, then the size of hidden positive pairs in each order can be boosted and the extracted group features will contain different order knowledge, resulting in a more regular embedding space. Specifically, as can be seen in Figure.\ref{Contrastive}, we randomly generate $q=10$ group features through MSRU from different orders of image group $\mathcal{I}$. Samples $z^u$ and $z^v$ are generated from the different orders of the same image group, and their corresponding memory banks are $\textbf{M}^u$ and $\textbf{M}^v$:
\begin{equation}
\textbf{M}^u = \{ m_i^u \}_{i \in {N}}, \textbf{M}^v = \{ m_i^v \}_{i \in {N}},
\end{equation}
where $N$ is the total number of different orders. The contrastive context of sample $z^u$ is the similarity set $S^u$ among $z^u$ and $\textbf{M}^u$, and we use relation miner $\Psi$ that generates $N_{+}$ positive samples:
\begin{equation}
\begin{split}
\textbf{S}^u &= \{ s_i^u \}_{i \in N} = \{ z^u \cdot m_i^u \}_{i \in N} \\ 
(\textbf{S}^u_{+},N_{+}) &= \Psi(\textbf{S}^u) = TopK(\textbf{S}^u)
\end{split}
\end{equation}
if we want to use the knowledge of order $u$ to guide order $v$ contrastive learning, it contains two aspects: 1): we select the most similar pairs (positive) in order $u$ as the positive sets in order $v$, i.e., $\textbf{S}^v$, $N_{+}^u \to \mathcal{C}(\textbf{S}^v|N_{+}^u)$. Thus the sample $z^v$ shares the positive neighbors of $z^u$. 2): we do not directly pull close the group feature embeddings generated from different orders, which would cause the collapse problem. We pull close the similarity of different samples. If sample $x$ is the positive sample of $z^u$, so $x$ would also be the positive sample of $z^v$. And we can calculate the similarities between \{$z^u$,$x$\} and \{$z^v$,$x$\}. Then we pull close these two similarities. The overall loss is conducted as:
\begin{equation}
L_{u \rightarrow v}=-\log \frac{\sum_{i \in N_{+}^{u}} \exp \left(s_{i}^{u} s_{i}^{v}\right) / \tau}{\sum_{i \in N} \exp \left(s_{i}^{u} s_{i}^{v}\right) / \tau}.
\end{equation}
Finally, the total contrastive loss can be written as:
\begin{equation}
L_{cocl} = L_{u \rightarrow v} + L_{v \rightarrow u}.
\end{equation}

Noting that all parts of our proposed network are trained jointly, so the total loss function is written as:
\begin{equation}
 L = L_{cocl} + L_{co}.
\end{equation}

\begin{table*}[!t]
\centering
\caption{Quantitative comparison with SOTA on five CoSOD datasets. The best two results are in {\color[HTML]{FF0000} red} and {\color[HTML]{32CB00} green}. Larger $E_\xi$, $S_m$, $F_\beta$, smaller MAE mean better results. ``*" means that the code or results are not available.}
\scalebox{0.72}{
\begin{tabular}{c|c|cccc|cccc|cccc|cccc|cccc}
\hline
                         &                        & \multicolumn{4}{c|}{CoCA}                                                                                                 & \multicolumn{4}{c|}{CoSOD3k}                                                                                              & \multicolumn{4}{c|}{Cosal2015}                                                                                            & \multicolumn{4}{c|}{iCoSeg}                                                                                               & \multicolumn{4}{c}{MSRC}                                                                                                  \\ \cline{3-22} 
\multirow{-2}{*}{Models} & \multirow{-2}{*}{Type} & $E_\xi$                      & $S_m$                        & $F_\beta$                    & MAE                          & $E_\xi$                      & $S_m$                        & $F_\beta$                    & MAE                          & $E_\xi$                      & $S_m$                        & $F_\beta$                    & MAE                          & $E_\xi$                      & $S_m$                        & $F_\beta$                    & MAE                          & $E_\xi$                      & $S_m$                        & $F_\beta$                    & MAE                          \\ \hline
CBCS(TIP2013)            & Co                     & 0.641                        & 0.523                        & 0.313                        & 0.180                        & 0.637                        & 0.528                        & 0.466                        & 0.228                        & 0.656                        & 0.544                        & 0.532                        & 0.233                        & 0.797                        & 0.658                        & 0.705                        & 0.172                        & 0.676                        & 0.480                        & 0.630                        & 0.314                        \\
GWD(IJCAI2017)           & Co                     & 0.701                        & 0.602                        & 0.408                        & 0.166                        & 0.777                        & 0.716                        & 0.649                        & 0.147                        & 0.802                        & 0.744                        & 0.706                        & 0.148                        & 0.841                        & 0.801                        & 0.829                        & 0.132                        & 0.789                        & 0.719                        & 0.727                        & 0.210                        \\
RCAN(IJCAI2019)          & Co                     & 0.702                        & 0.616                        & 0.422                        & 0.160                        & 0.808                        & 0.744                        & 0.688                        & 0.130                        & 0.842                        & 0.779                        & 0.764                        & 0.126                        & 0.878                        & 0.820                        & 0.841                        & 0.122                        & 0.789                        & 0.719                        & 0.727                        & 0.210                        \\
CSMG(CVPR2019)           & Co                     & 0.735                        & 0.632                        & 0.508                        & 0.124                        & 0.804                        & 0.711                        & 0.709                        & 0.157                        & 0.842                        & 0.774                        & 0.784                        & 0.130                        & 0.889                        & 0.821                        & 0.850                        & 0.106                        & 0.859                        & 0.722                        & 0.847                        & 0.190                        \\
CoEG(TPAMI2020)          & Co                     & 0.717                        & 0.616                        & 0.499                        & {\color[HTML]{00B050} 0.104} & 0.825                        & 0.762                        & 0.736                        & 0.092                        & 0.882                        & 0.836                        & 0.832                        & 0.077                        & 0.912                        & 0.875                        & 0.876                        & 0.060                        & 0.793                        & 0.696                        & 0.751                        & 0.188                        \\
GICD(ECCV2020)           & Co                     & 0.712                        & 0.658                        & 0.510                        & 0.125                        & 0.831                        & 0.778                        & 0.744                        & 0.089                        & 0.885                        & 0.842                        & 0.840                        & 0.071                        & 0.891                        & 0.832                        & 0.845                        & 0.068                        & 0.726                        & 0.665                        & 0.692                        & 0.196                        \\
ICNet(NeurIPS2020)          & Co                     & 0.698                        & 0.651                        & 0.506                        & 0.148                        & 0.832                        & 0.780                        & 0.743                        & 0.097                        & 0.900                        & 0.856                        & 0.855                        & {\color[HTML]{00B050} 0.058} & 0.929                        & 0.869                        & 0.886                        & 0.047                        & 0.822                        & 0.731                        & 0.805                        & 0.160                        \\
CoADNet(NeurIPS2020)           & Co                     & *                            & *                            & *                            & *                            & {\color[HTML]{00B050} 0.874} & {\color[HTML]{00B050} 0.822} & {\color[HTML]{00B050} 0.786} & 0.078                        & {\color[HTML]{333333} 0.915} & {\color[HTML]{333333} 0.861} & {\color[HTML]{333333} 0.857} & {\color[HTML]{333333} 0.063} & {\color[HTML]{00B050} 0.930} & {\color[HTML]{00B050} 0.878} & {\color[HTML]{333333} 0.889} & {\color[HTML]{00B050} 0.045} & {\color[HTML]{333333} 0.850} & {\color[HTML]{333333} 0.782} & {\color[HTML]{333333} 0.842} & {\color[HTML]{333333} 0.132} \\
GCoNet(CVPR2021)         & Co                     & {\color[HTML]{00B050} 0.760} & {\color[HTML]{333333} 0.673} & {\color[HTML]{333333} 0.544} & 0.105                        & 0.860                        & 0.802                        & 0.777                        & {\color[HTML]{00B050} 0.071} & 0.888                        & 0.845                        & 0.847                        & 0.068                        & 0.886                        & 0.834                        & 0.839                        & 0.068                        & 0.736                        & 0.663                        & 0.715                        & 0.188                        \\
CADC(ICCV2021)           & Co                     & {\color[HTML]{333333} 0.744} & {\color[HTML]{00B050} 0.681} & {\color[HTML]{00B050} 0.548} & {\color[HTML]{333333} 0.132} & {\color[HTML]{333333} 0.840} & {\color[HTML]{333333} 0.801} & {\color[HTML]{333333} 0.759} & {\color[HTML]{333333} 0.096} & 0.906                        & {\color[HTML]{00B050} 0.866} & 0.862                        & 0.064                        & 0.910                        & 0.868                        & 0.856                        & 0.063                        & {\color[HTML]{FF0000} 0.895} & {\color[HTML]{00B050} 0.821} & {\color[HTML]{FF0000} 0.873} & {\color[HTML]{FF0000} 0.115} \\
GLNet(TCyb2022)          & Co                     & 0.716                        & 0.591                        & 0.441                        & 0.188                        & *                            & *                            & *                            & *                            & {\color[HTML]{00B050} 0.925} & 0.855                        & {\color[HTML]{00B050} 0.885} & {\color[HTML]{333333} 0.060} & {\color[HTML]{00B050} 0.930} & {\color[HTML]{333333} 0.874} & {\color[HTML]{00B050} 0.899} & {\color[HTML]{00B050} 0.045} & {\color[HTML]{00B050} 0.890} & {\color[HTML]{FF0000} 0.830} & 0.869                        & {\color[HTML]{00B050} 0.120} \\ \hline \hline
Ours(32)                 & Co                     & {\color[HTML]{FF0000} 0.776} & {\color[HTML]{FF0000} 0.732} & {\color[HTML]{FF0000} 0.616} & {\color[HTML]{FF0000} 0.099} & {\color[HTML]{FF0000} 0.888} & {\color[HTML]{FF0000} 0.843} & {\color[HTML]{FF0000} 0.820} & {\color[HTML]{FF0000} 0.062} & {\color[HTML]{FF0000} 0.934} & {\color[HTML]{FF0000} 0.898} & {\color[HTML]{FF0000} 0.902} & {\color[HTML]{FF0000} 0.045} & {\color[HTML]{FF0000} 0.948} & {\color[HTML]{FF0000} 0.911} & {\color[HTML]{FF0000} 0.916} & {\color[HTML]{FF0000} 0.035} & {\color[HTML]{333333} 0.881} & {\color[HTML]{333333} 0.809} & {\color[HTML]{00B050} 0.870} & {\color[HTML]{FF0000} 0.115} \\
Ours(16)                 & Co                     & 0.770                        & 0.727                        & 0.608                        & 0.103                        & 0.883                        & 0.839                        & 0.815                        & 0.065                        & 0.929                        & 0.896                        & 0.897                        & 0.047                        & 0.942                        & 0.905                        & 0.907                        & 0.039                        & 0.875                        & 0.801                        & 0.865                        & 0.121                        \\
Ours(8)                  & Co                     & 0.765                        & 0.722                        & 0.603                        & 0.106                        & 0.879                        & 0.836                        & 0.811                        & 0.067                        & 0.924                        & 0.893                        & 0.893                        & 0.050                        & 0.937                        & 0.900                        & 0.901                        & 0.043                        & 0.871                        & 0.797                        & 0.859                        & 0.123                        \\ \hline \hline
EGNet(ICCV2019)          & Sin                    & 0.631                        & 0.595                        & 0.388                        & 0.179                        & 0.793                        & 0.762                        & 0.702                        & 0.119                        & 0.843                        & 0.818                        & 0.786                        & 0.099                        & 0.911                        & 0.875                        & 0.875                        & 0.060                        & 0.794                        & 0.702                        & 0.752                        & 0.186                        \\
F3Net(AAAI2020)          & Sin                    & 0.678                        & 0.614                        & 0.437                        & 0.178                        & 0.802                        & 0.772                        & 0.717                        & 0.114                        & 0.866                        & 0.841                        & 0.815                        & 0.084                        & 0.918                        & 0.879                        & 0.874                        & 0.048                        & 0.811                        & 0.733                        & 0.763                        & 0.161                        \\
MINet(CVPR2020)          & Sin                    & 0.634                        & 0.550                        & 0.387                        & 0.221                        & 0.782                        & 0.754                        & 0.707                        & 0.122                        & 0.847                        & 0.831                        & 0.805                        & 0.181                        & 0.846                        & 0.789                        & 0.784                        & 0.099                        & 0.769                        & 0.688                        & 0.729                        & 0.194                        \\ \hline
\end{tabular}}
\label{cosod_QuantitativeResults1}
\vspace{-0.3cm}
\end{table*}

\begin{figure*}[!t]
\centering
\includegraphics[scale=0.32]{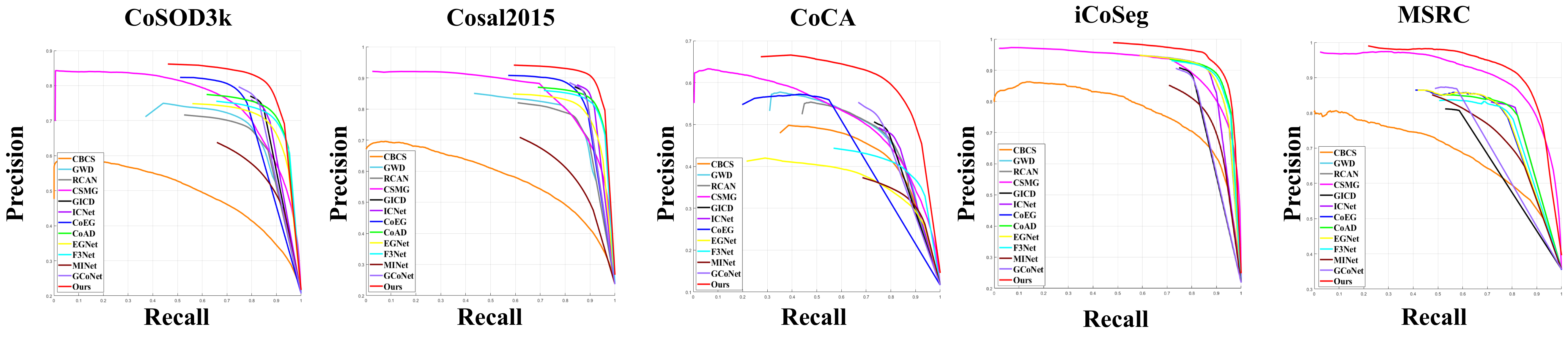}
\caption{Comparison of PR curves across five CoSOD datasets.}
\label{cosod_pr}
\end{figure*}

\begin{table}[t!]
\begin{minipage}{0.48\linewidth}
\centering
\caption{Results on Internet Dataset.}
\scalebox{0.8}{
\begin{tabular}{c|cccc}
\hline
\multicolumn{1}{l|}{} & Airplane                     & Car                          & Horse                        & $Avg.J$                      \\ \hline
Li19                  & 0.830                        & 0.930                        & 0.760                        & 0.840                        \\
Li21                  & {\color[HTML]{32CB00} 0.840} & {\color[HTML]{32CB00} 0.920} & {\color[HTML]{FF0000} 0.830} & {\color[HTML]{32CB00} 0.863} \\
Ours                  & {\color[HTML]{FF0000} 0.868} & {\color[HTML]{FF0000} 0.951} & {\color[HTML]{32CB00} 0.802} & {\color[HTML]{FF0000} 0.874} \\ \hline
\end{tabular}}
\label{internet}
\end{minipage}
\begin{minipage}{0.48\linewidth}\quad
\centering
\caption{Results on Pascal Dataset.}
\scalebox{0.8}{
\begin{tabular}{c|cc}
\hline
\multicolumn{1}{l|}{} & $P$                          & $J$                          \\ \hline
Li19                  & 0.940                        & 0.630                        \\
Li21                  & {\color[HTML]{32CB00} 0.970} & {\color[HTML]{32CB00} 0.740} \\
Ours                  & {\color[HTML]{FE0000} 0.973} & {\color[HTML]{FE0000} 0.746} \\ \hline
\end{tabular}}
\label{pascal}
\end{minipage}

\begin{minipage}{0.48\linewidth}
\centering
\caption{\centering Results on iCoseg Dataset.}
\scalebox{0.65}{
\begin{tabular}{c|ccccccccc}
\hline
\multicolumn{1}{l|}{} & bear2                        & brownbear                    & cheetah                      & elephant                     & helicopter                   & hotballoon                   & panda1                       & panda2                       & $Avg.J$                      \\ \hline
Li19                  & 0.901                        & 0.897                        & 0.920                        & 0.902                        & 0.760                        & 0.917                        & 0.902                        & 0.898                        & 0.887                        \\ \hline
Li21                  & {\color[HTML]{32CB00} 0.928} & {\color[HTML]{FE0000} 0.941} & {\color[HTML]{32CB00} 0.891} & {\color[HTML]{32CB00} 0.916} & {\color[HTML]{32CB00} 0.852} & {\color[HTML]{32CB00} 0.951} & {\color[HTML]{32CB00} 0.948} & {\color[HTML]{FE0000} 0.921} & {\color[HTML]{32CB00} 0.921} \\ \hline
Ours                  & {\color[HTML]{FE0000} 0.933} & {\color[HTML]{32CB00} 0.934} & {\color[HTML]{FE0000} 0.926} & {\color[HTML]{FE0000} 0.932} & {\color[HTML]{FE0000} 0.883} & {\color[HTML]{FE0000} 0.964} & {\color[HTML]{FE0000} 0.949} & {\color[HTML]{32CB00} 0.902} & {\color[HTML]{FE0000} 0.927} \\ \hline
\end{tabular}}
\label{icoseg}
\end{minipage}
\end{table}

\section{Experimental Results} \label{experimental}
\subsection{Implementation Details}
{  Most of the previous methods~\cite{DBLP:conf/cvpr/ZhangLSLC020,jin2020icnet,GLNet} use the VGG16 as the backbone. Therefore, for a fair comparison, we also choose the VGG16 network to extract the features of each image in the group, which is the dominant reason. Moreover, the CoSOD task deals with image group data, and the core problem in CoSOD is group-feature learning, which cannot be solved by simply replacing the stronger backbones. The same conclusion is reported in the work GLNet~\cite{GLNet}.} The training set is a subset of the COCO dataset~\cite{DBLP:conf/eccv/LinMBHPRDZ14} (9213 images) and saliency dataset DUTS~\cite{DBLP:conf/cvpr/WangLWF0YR17}, as suggested by~\cite{DBLP:conf/nips/ZhangCHLZ20}. All the images are resized to the same size of $352 \times 352$ for easy processing. The model is optimized by the Adam algorithm with a weight decay of 5e-4 and an initial learning rate of 1e-4. During training, the batchsize is 32. For co-saliency detection, the training of our proposed network includes two stages: 

\textit{Stage1.} We first train our model using DUTS
dataset~\cite{DBLP:conf/cvpr/WangLWF0YR17} to focus on the salient areas. Note that when training, to match the size of input group, we augment the single salient image to $N=32$ different images as a group using affine transformation, horizontal flipping and left-right flipping.

\textit{Stage2.} We further fine-tune our model using sub-coco dataset to better focus on the co-salient areas. All the parameter settings are the same as those in \textit{Stage1}. 

As described, since CoSEG dose not need to detect these objects belong to salient regions, we only use \textit{Stage2} when training CoSEG model.

\subsection{Evaluation Datasets and Metrics}
\textbf{Co-saliency Detection.} We employ five challenging datasets for evaluation: CoCA~\cite{DBLP:conf/eccv/ZhangJXC20}, CoSOD3k~\cite{deng2020re}, Cosal2015~\cite{DBLP:journals/ijcv/ZhangHLWL16}, iCoseg~\cite{DBLP:conf/cvpr/BatraKPLC10} and MSRC~\cite{DBLP:conf/iccv/WinnCM05}. Cosal2015 has 50 groups and a total of 2015 images. Cosal2015 suffers from various challenging factors such as complex environments, occlusion issues, target appearance variations and background clutters, etc. CoSOD3k~\cite{deng2020re} has 160 groups and a total of 3000 images. CoSOD3k is the largest-scale and most comprehensive benchmark, which has sufficient object diversity and the complexity for size and number for instances. CoCA has 80 groups and a total of 1297 images. CoCA is a challenging dataset, since the images typically contain other multiple objects in addition to the co-salient objects which are even smaller in size. iCoseg consists of 38 groups of total 643 images. MSRC contains 7 groups of total 240 images, and each group has $30 \sim 53$ images.

To evaluate the performance of the proposed method, we adopted five widely used criteria: (1) Precision-Recall (PR) curve, which shows the tradeoff between precision and recall for different threshold (ranging from 0 to 255). (2) The F-measure ($F_\beta$), which denotes the harmonic mean of the precision and recall values obtained by a self-adaptive threshold $T = \mu +\sigma$ ($\mu$ and $\sigma$ are the mean value and standard deviation of co-saliency map):
\begin{equation}
    F_{\beta} = \frac{(1+\beta^{2})\times Precision \times Recall}{\beta^{2}Precision+Recall },
\end{equation}
where $\beta^{2}$ is typically set to 0.3 as suggested in~\cite{DBLP:journals/tcsv/HanCLZ18,DBLP:journals/tip/BorjiCJL15, DBLP:conf/cvpr/YangZLRY13}. 
In this paper, we use maximum F-measure $F_\beta$ to evluate the performance. (3) Structure Measure ($S_{m}$) is adopted to evaluate the spatial structure similarities of saliency maps based on both region-aware structural similarity $S_{r}$ and object-aware structural similarity $S_{o}$, defined as
\begin{equation}
    S_{m} = \alpha \ast S_{r} + (1- \alpha) \ast S_{o},
\end{equation}
where $\alpha = 0.5$~\cite{DBLP:conf/iccv/FanCLLB17}. (4) The E-measure~\cite{DBLP:conf/ijcai/FanGCRCB18} is a perceptual metric that evaluates both local and global similarity between the predicted map and ground-truth simultaneously. In this paper, we use maximum E-measure $E_{\xi}$. (5). Mean absolute error (MAE), which characterize the average 1-norm distance between ground truth maps and predictions. Evaluation toolbox:\url{ https://dpfan.net/CoSOD3K/}.

\begin{figure*}[]
\centering
\includegraphics[scale=0.55]{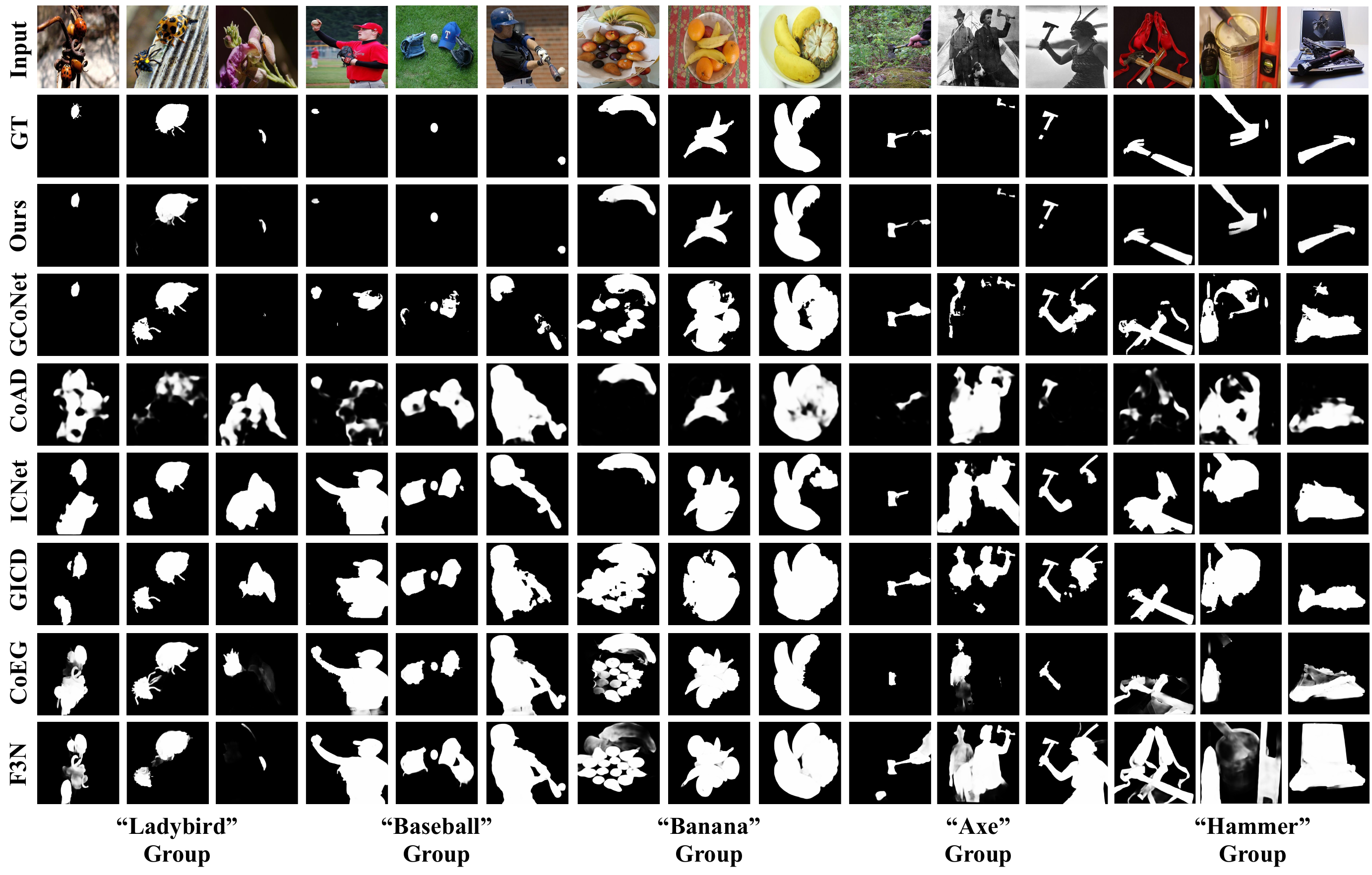}
\caption{Qualitative Evaluation on CoSOD task.}
\label{cosod_QualitativeResults}
\end{figure*}

\textbf{Object Co-segmentation.} We evaluate the proposed method and compare it with existing methods on three benchmarks for object co-segmentation, including the Internet dataset~\cite{DBLP:conf/cvpr/RubinsteinJKL13}, the iCoseg dataset~\cite{DBLP:conf/cvpr/BatraKPLC10}, and the PASCAL-VOC dataset~\cite{DBLP:conf/iccv/FaktorI13}. These datasets are composed of real-world images with large intra-class variations, occlusions and background clutters. The Internet dataset contains images of three object categories including airplane, car and horse. Thousands of images in this dataset were collected from the Internet. Following the same setting of the previous work~\cite{DBLP:conf/ijcai/HsuLC18,DBLP:conf/cvpr/RubinsteinJKL13,DBLP:conf/aaai/TaoLFF17} , we use the same subset of the Internet dataset where 100 images per class are available. 
iCoseg consists of 38 groups of total 643 images which are challenging for object co-segmentation task because of the large variations of viewpoints and multiple co-occurring object instances. The PASCAL-VOC dataset contains total 1,037 images of 20 object classes from PASCAL-VOC 2010 dataset. The PASCAL-VOC dataset is more challenging and difficult than the Internet dataset due to extremely large intra-class variations and subtle figure-ground discrimination. Two widely used measures, \textit{precision} ($\mathcal{P}$) and \textit{Jaccard index} ($\mathcal{J}$), are adapted to evaluate the performance of object co-segmentation. Precision measures the percentage of correctly segmented pixels including both object and background pixels. Jaccard index is the ratio of the intersection area of the detected objects and the ground truth to their union area. 

\begin{table}[!t]
\centering
\caption{ { Efficiency comparison with some SOTA methods.}}
\scalebox{1.0}{
\begin{tabular}{c|ccc}
\hline
Models & Year        & Model Size(MB) & FPS \\ \hline
GICD   & ECCV2020    & 278.04      & 55  \\
ICNet  & NeurIPS2020 & 70.41       & 80  \\
CoADNet   & NeurIPS2020 & 289.23      & 14  \\
GCoNet & CVPR2021    & 540.36      & 59  \\
CADC   & ICCV2021    & 392.85      & 15  \\
GLNet  & TCYB2022    & 237.12      & 35  \\
GCAGC  & CVPR2020    & 281.81      & 25  \\
RCAN   & IJCAI2019   & 150.34      & 54  \\
Ours   & *           & 208.23      & 27  \\ \hline
\end{tabular}}
\label{Efficiency}
\end{table}

\subsection{Comparisons with the State-of-the-Arts}

{For CoSOD task, we compare our approach with 11 CoSOD models and 3 SOD models: CBCS~\cite{DBLP:journals/tip/FuCT13}, GWD~\cite{DBLP:journals/tip/WeiZBLWZ19}, RCAN~\cite{DBLP:conf/ijcai/0061STSS19}, CSMG~\cite{DBLP:conf/cvpr/ZhangLL019}, CoEG~\cite{9358006}, GICD~\cite{DBLP:conf/eccv/ZhangJXC20}, ICNet~\cite{jin2020icnet}, CoADNet~\cite{DBLP:conf/nips/ZhangCHLZ20}, GCoNet~\cite{DBLP:conf/cvpr/FanFFT0T21},
CADC~\cite{DBLP:conf/iccv/ZhangHL021}, GLNet~\cite{GLNet}, EGNet~\cite{DBLP:conf/iccv/ZhaoLFCYC19}, F3Net~\cite{DBLP:journals/corr/abs-1911-11445} and MINet~\cite{DBLP:conf/cvpr/PangZZL20}.} For CoSEG task, we compare our approach with other 2 most SOTA methods: Li19~\cite{DBLP:conf/iccv/LiSLWH19} and Li21~\cite{DBLP:journals/tip/ZhangLLWY21}.

\begin{figure*}[]
\centering
\includegraphics[scale=0.5]{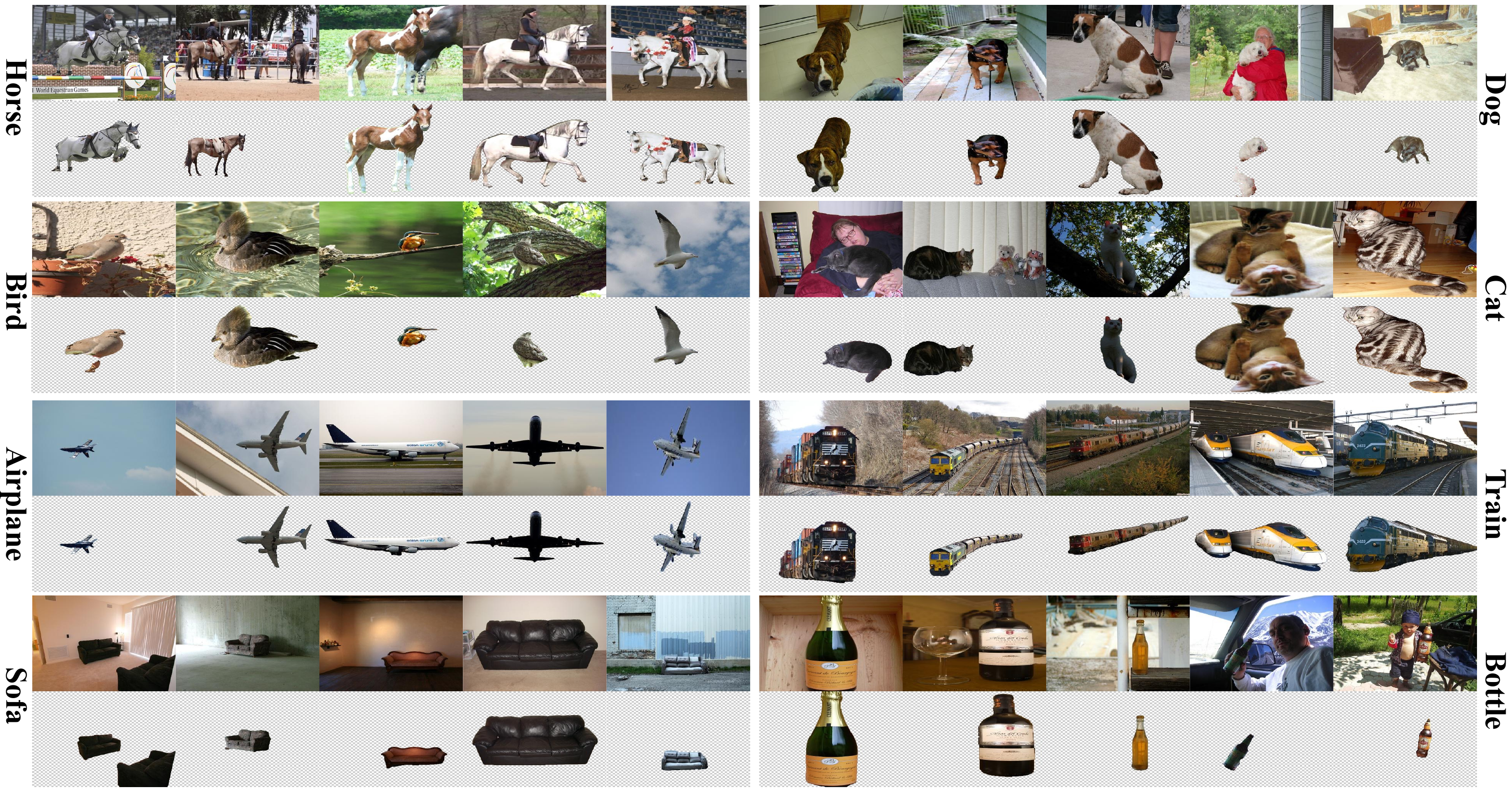}
\caption{Qualitative Evaluation on CoSEG task.}
\label{coseg_QualitativeResults}
\end{figure*}

\begin{table*}[!htbp]
\centering
\caption{Architecture ablation studies}
\scalebox{0.82}{
\begin{tabular}{l|cccc|cccc|cc}
\hline
\multicolumn{1}{c|}{\multirow{2}{*}{Configurations}} & \multicolumn{4}{c|}{CoSOD3k}        & \multicolumn{4}{c|}{Cosal2015}      & \multicolumn{2}{c}{PASCAL-VOC} \\ \cline{2-11} 
\multicolumn{1}{c|}{}                                & $E_\xi$ & $S_m$ & $F_\beta$ & MAE   & $E_\xi$ & $S_m$ & $F_\beta$ & MAE   & $P$            & $J$           \\ \hline
1. Baseline                                          & 0.788   & 0.718 & 0.660     & 0.139 & 0.810   & 0.750 & 0.714     & 0.140 & 0.891          & 0.575         \\
2. Baseline+GRU                                      & 0.859   & 0.820 & 0.795     & 0.075 & 0.914   & 0.872 & 0.874     & 0.061 & 0.927          & 0.702         \\
3. Baseline+LSTM                                     & 0.861   & 0.819 & 0.796     & 0.073 & 0.913   & 0.875 & 0.875     & 0.059 & 0.930          & 0.701         \\
4. Baseline+RCAU                                     & 0.866   & 0.826 & 0.800     & 0.071 & 0.916   & 0.878 & 0.884     & 0.054 & 0.942          & 0.720         \\
5. Baseline+MSRU                                     & 0.876   & 0.833 & 0.810     & 0.066 & 0.924   & 0.887 & 0.890     & 0.049 & 0.958          & 0.732         \\
6. Baseline+MSRU+NLFM                                & 0.880   & 0.836 & 0.814     & 0.065 & 0.928   & 0.890 & 0.896     & 0.048 & 0.962          & 0.737         \\
7. Baseline+MSRU+NLFM+$L_{cont}$(Ours)               & 0.888   & 0.843 & 0.820     & 0.062 & 0.934   & 0.898 & 0.902     & 0.045 & 0.973          & 0.746         \\ \hline
8. Baseline+NLFM+$L_{cont}$+RCAU                     & 0.872   & 0.829 & 0.806     & 0.067 & 0.923   & 0.885 & 0.888     & 0.052 & 0.958          & 0.728         \\
9. Baseline+NLFM+$L_{cont}$+RCAU+DOM                 & 0.878   & 0.836 & 0.811     & 0.065 & 0.926   & 0.890 & 0.895     & 0.050 & 0.964          & 0.730         \\
10. Baseline+NLFM+$L_{cont}$+RU                      & 0.879   & 0.837 & 0.811     & 0.066 & 0.926   & 0.891 & 0.896     & 0.050 & 0.963          & 0.729         \\
11.   Baseline+NLFM+$L_{cont}$+DOM+RU(-NLCA)         & 0.883   & 0.839 & 0.815     & 0.065 & 0.929   & 0.893 & 0.896     & 0.048 & 0.967          & 0.736         \\
12.   Baseline+NLFM+$L_{cont}$+DOM+RU(-CFPM)         & 0.882   & 0.840 & 0.814     & 0.064 & 0.928   & 0.892 & 0.895     & 0.049 & 0.966          & 0.738         \\ \hline
13. Ours($k=4$)                                      & 0.890   & 0.843 & 0.822     & 0.061 & 0.937   & 0.897 & 0.904     & 0.043 & 0.974          & 0.748         \\
14. Ours($k=5$)                                      & 0.889   & 0.844 & 0.821     & 0.060 & 0.935   & 0.898 & 0.903     & 0.044 & 0.973          & 0.745         \\ \hline
15. Baseline+MSRU+$L_{cont}$(8+noise)                & 0.877   & 0.835 & 0.816     & 0.065 & 0.927   & 0.894 & 0.896     & 0.049 & 0.968          & 0.738         \\
16. Baseline+MSRU+$L_{cont}$(16+noise)               & 0.884   & 0.840 & 0.819     & 0.063 & 0.931   & 0.896 & 0.899     & 0.046 & 0.970          & 0.741         \\
17. Baseline+MSRU+$L_{cont}$(32+noise)               & 0.886   & 0.841 & 0.820     & 0.063 & 0.932   & 0.897 & 0.901     & 0.046 & 0.972          & 0.745         \\ \hline
\end{tabular}}
\label{ablation_art}
\end{table*}

\begin{table*}[]
\centering
\caption{Stability ablation studies.}
\scalebox{0.65}{
\begin{tabular}{c|cccc|cccc|cc}
\hline
\multirow{2}{*}{Configurations}       & \multicolumn{4}{c|}{CoSOD3k}                                                  & \multicolumn{4}{c|}{Cosal2015}                                                  & \multicolumn{2}{c}{PASCAL-VOC}        \\ \cline{2-11} 
                                      & $E_\xi$           & $S_m$             & $F_\beta$         & MAE               & $E_\xi$           & $S_m$             & $F_\beta$           & MAE               & $P$               & $J$               \\ \hline
Ours(RU+DOM+$L_{cocl}$)               & $0.887 \pm 0.004$ & $0.842 \pm 0.003$ & $0.818 \pm 0.004$ & $0.060 \pm 0.002$ & $0.934 \pm 0.002$ & $0.897 \pm 0.004$ & $ 0.899 \pm 0.005$  & $0.041 \pm 0.003$ & $0.974 \pm 0.003$ & $0.747 \pm 0.002$ \\
\textbf{Ours(RU+DOM+\bm{$L_{cocl}$}, std)} & \textbf{0.002}    & \textbf{0.002}    & \textbf{0.003}    & \textbf{0.001}    & \textbf{0.001}    & \textbf{0.003}    & \textbf{0.004}      & \textbf{0.002}    & \textbf{0.002}    & \textbf{0.001}    \\ \hline
Ours(RU+$L_{cocl}$)                   & $0.875 \pm 0.007$ & $0.836 \pm 0.006$ & $0.809 \pm 0.008$ & $0.064 \pm 0.005$ & $0.922 \pm 0.008$ & $0.885 \pm 0.010$ & $ 0.892 \pm 0.008$  & $0.042 \pm 0.006$ & $0.961 \pm 0.006$ & $0.731 \pm 0.008$ \\
\textbf{Ours(RU+\bm{$L_{cocl}$}, std)}     & \textbf{0.006}    & \textbf{0.005}    & \textbf{0.007}    & \textbf{0.004}    & \textbf{0.007}    & \textbf{0.008}    & \textbf{0.006}      & \textbf{0.004}    & \textbf{0.005}    & \textbf{0.006}    \\ \hline
Ours(RU)                            & $0.865 \pm 0.010$ & $0.822 \pm 0.013$ & $0.796 \pm 0.012$ & $0.075 \pm 0.010$ & $0.913 \pm 0.011$ & $0.876 \pm 0.012$ & $ 0.884 \pm 0.011$  & $0.059 \pm 0.012$ & $0.952 \pm 0.012$ & $0.722 \pm 0.011$ \\
\textbf{Ours(RU,std)}              & \textbf{0.009}    & \textbf{0.011}    & \textbf{0.011}    & \textbf{0.009}    & \textbf{0.010}    & \textbf{0.010}    & \textbf{0.010}      & \textbf{0.011}    & \textbf{0.010}    & \textbf{0.009}    \\ \hline
RCAN                                  & $0.796 \pm 0.012$ & $0.734 \pm 0.010$ & $0.679 \pm 0.009$ & $0.143 \pm 0.013$ & $0.829 \pm 0.013$ & $0.766 \pm 0.013$ & $ 0.753 \pm 0.011 $ & $0.138 \pm 0.012$ & $0.920 \pm 0.010$ & $0.620 \pm 0.010$ \\
\textbf{RCAN(Std)}                    & \textbf{0.011}    & \textbf{0.009}    & \textbf{0.008}    & \textbf{0.010}    & \textbf{0.011}    & \textbf{0.011}    & \textbf{0.009}      & \textbf{0.011}    & \textbf{0.009}    & \textbf{0.009}    \\ \hline
ICNet                                 & $0.821 \pm 0.011$ & $0.769 \pm 0.011$ & $0.733 \pm 0.010$ & $0.109 \pm 0.012$ & $0.889 \pm 0.011$ & $0.846 \pm 0.010$ & $ 0.843 \pm 0.012$  & $0.068 \pm 0.010$ & *                 & *                 \\
\textbf{ICNet(Std)}                   & \textbf{0.010}    & \textbf{0.010}    & \textbf{0.009}    & \textbf{0.011}    & \textbf{0.010}    & \textbf{0.010}    & \textbf{0.010}      & \textbf{0.009}    & \textbf{*}        & \textbf{*}        \\ \hline
\end{tabular}}
\label{Stability}
\end{table*}

\textbf{Quantitative Evaluation.}
From Table.\ref{cosod_QuantitativeResults1}, we can see that compared to other state-of-the-art methods, our model (Ours(32)) outperforms all of other SOTA methods in all metrics. As reported in Table.\ref{cosod_QuantitativeResults1}, our approach achieves good performance on different size groups (8, 16 and 32) and still consistently outperforms all the state-of-the-art methods. And the performance raises along with the group size, which emphasizes the importance of the group information completeness to robust co-saliency detection. { In dataset CoCA, compared to the second ranked performance, the improvement of our proposed method reaches 2.1\% for $E_\xi$,  7.4\% for $S_m$, 12.4\% for $F_\beta$ and 5.7\% for MAE. } On the challenging CoSOD3k and Cosal2015 datasets, our model capitalizes on our better consensus and significantly outperforms other methods. These results demonstrate the efficiency of the proposed framework. As shown in Fig.\ref{cosod_pr}, we can see that our method (the red line) achieves the highest precision on all datasets. Table.\ref{internet}, Table.\ref{pascal} and Table.\ref{icoseg} show the results on CoSEG datasets, it can be seen that our method can outperform all other two SOTA methods in most datasets. { Finally, we show the parameter number and running time comparison with other SOTA methods, and the results are shown in the Table.\ref{Efficiency}.  Our proposed model runs at a competitive efficiency compared to other models. This is primarily due to the fact that cross-order contrastive loss is only used during training, and only the dummy orders mechanism (DOM) would slow the network down to some extent.}

\textbf{Qualitative Evaluation.}
Fig.\ref{cosod_QualitativeResults} shows the co-saliency maps generated by different methods for qualitative comparison. As can be seen, the SOD method F3N can only detect salient objects and fail to distinguish co-salient objects. The CoSOD methods perform better than the SOD methods because of considering group-wise relationships in designing the model. As can be seen in "Baseball Group", these CoSOD can suppress some non-co-salient regions. However, these CoSOD methods require constant input data, or only using simple adding operation to generate final group features. When facing complex real-world scenarios, they are unable to handle these challenging cases, like "Hammer Group" and "Axe Group".  While our proposed model can capture complete inter-saliency relations of an image-group, and make the training and inference process more stable, therefore performs much better on detecting co-salient objects. Fig.\ref{coseg_QualitativeResults} shows the co-segmentation maps generated by our proposed methods for qualitative comparison, which can further demonstrate the superiority of our model in CoSEG task.  

\subsection{Ablation Studies}
In this section, we first conduct evaluation on CoSOD3K, Cosal2015 and PASCAL-VOC datasets to investigate the effectiveness of various components of the proposed model. We first do architecture ablation studies, and we set the baseline model by only using single feature learning branch and replacing the recurrent neural network with concatenation operation. \textbf{The baseline model is only trained with co-saliency loss $L_{co}$ alone. Moreover, we set the input orders of different models the same, so the performance changes of different models are a result of different architectures.} The results are shown in Table.\ref{ablation_art}.  Secondly, we further evaluate the stability of our proposed network, and the results are shown in Table.\ref{Stability}.

\textbf{Architecture ablation studies.} As can be seen in No.1 of Table.\ref{ablation_art}, even though trained with co-saliency loss $L_{co}$, the baseline model can not well handle the co-saliency or co-segmentation task. When we replace the simple concatenation operation with typical recurrent unit, GRU or LSTM (No.2 and No.3), the performance is improved compared to baseline, which means the recurrent architecture is suitable for co-saliency detection task. Moreover, we use our previous RCAU to replace GRU or LSTM (No.4), the performance can be further improved, because proposed RCAU can generate more robust group features compared to GRU or LSTM. 
Then, we add our proposed MSRU (No.5), the performance has large improvement. This is because our proposed MSRU addresses the order-sensitive problem of RCAU and further improve the stability of proposed network. It should be noted that in this paper, we add GRU, LSTM, RCAU and MSRU on three levels of
backbone network (Conv4-3, Conv5-3 and Conv6-3), which can help capture multi-level inter-saliency relations. Then, adding the NLFM (No.6) can further improve the performance. Finally, COCL (No.7) can further boost the performance because it can help pull close group feature embeddings generated from different orders. To further justify the denoising ability of our MSRU, we add a noise image which is randomly selected from COCO dataset to the testing groups in size 8, 16 and 32 (No.15, No.16 and No.17). As shown in results, although the noise data damages our performance a little, we still outperforms all the SOTA methods, which demonstrate the robustness of the proposed method.

We also investigate the effectiveness our proposed DOM, and NLCA or CFPM in RU. Compared No.9 to No.8, It can be seen that when adding DOM on RCAU, it can improve the performance. Compared No.10 to No.7, when removing the DOM from our proposed method, the performance would be declined. It shows the effectiveness our proposed DOM. As can be seen in No.13 and No.14, when we set window size $k=4$ or $k=5$ in DOM (Eq.3), the performance has no obvious change while the training/testing is increasing about 30\%. So in this paper, we only set $k=3$. When we remove the reset gate $g_{d}$ from MSRU (No.11), the performance declines on three metrics especially on MAE. This indicates the reset gate $g_{d}$ is able to suppress the noise information in the group. When we remove the update gate $g_z$ from MSRU (No.12), the performance declines which indicates update gate $g_z$ can well retain group information in group feature and determine what new information should be updated from current  image feature representation. These experiments verify the effectiveness of the different modules proposed in this paper.

\begin{figure*}[!t]
\centering
\includegraphics[scale=0.4]{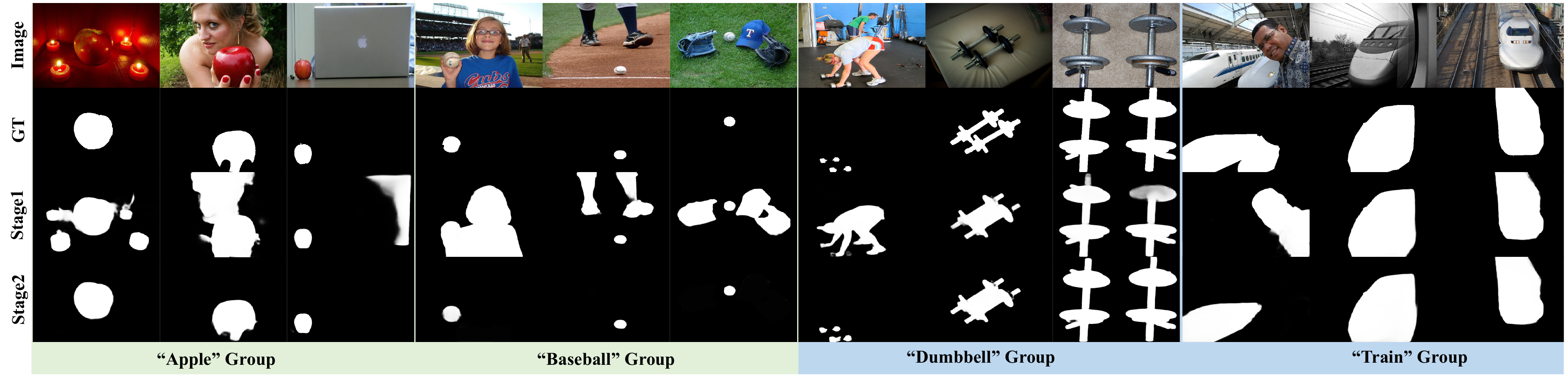}
\caption{ { Visualization comparison between stage1 and stage2 models.} }
\label{fig_stage1}
\end{figure*}

\textbf{Stability ablation studies.} During testing of our proposed method (Ours(RU+DOM+$L_{cocl}$)), for each image group, we randomize 10 different orders. From Table.\ref{Stability}, it can be seen that performance has no obvious change. \textbf{``Std"} means the standard deviation of these 10 orders.  This result verifies that our proposed MSRU and Contrastive Loss can let the proposed method less insensitive to the input order of group images. In the Introduction of this paper, we argue that existing sequential order modeling approaches make CoSOD networks unstable. So we do experiments on two typical methods, including RCAN~\cite{DBLP:conf/ijcai/0061STSS19} and ICNet~\cite{jin2020icnet}, to verify the inferring procedure of these two methods are unstable. Because CoADNet~\cite{DBLP:conf/nips/ZhangCHLZ20} does not release their code, so we can not do experiments on CoADNet. During testing, for each image group, we randomize 10 different orders, and the results are shown in Table.\ref{Stability}. As can be seen in Table.\ref{Stability}, whether using RNN (  RCAN(Std) ), or applying some sophisticated modification on CNNs architectures (  ICNet(Std) ), can not let the CoSOD network be insensitive to the input order of group images, leading to an unstable inferring procedure. Because in sequential order modeling, both CNNs and RNNs have inherent deficiencies. In this paper, our proposed DOM and COCL can help eliminate the effects of different orders. Therefore, as can be seen in Table.\ref{Stability}, when we remove these two modules, our proposed network would be more sensitive to the order of images input. Through these experiments, we further verify that our proposed network greatly improves the stability of CoSOD. \textbf{In Table.\ref{icoseg} and Table.\ref{cosod_QuantitativeResults1}, we test our network 10 times and select the medium performance in this paper. From the first row of the Table.\ref{Stability}, the worst performance of our proposed network remains SOTA.} {  Note that the FPS of the models Ours(RU) and Ours(RU+DOM) are 44 and 27. Compared to the previous RCAN, our proposed RU is slower because we use non-local attention. Moreover, adding DOM mechanism further reduces the running speed. However, as shown in the Table.\ref{Efficiency}, our proposed model still runs at a competitive efficiency compared to other SOTA models.}

\begin{table}[!t]
\centering
\caption{ {  Quantitative comparison between stage1 and stage2 models.}}
\scalebox{0.8}{
\begin{tabular}{c|cccc|cccc}
\hline
\multirow{2}{*}{Models} & \multicolumn{4}{c|}{CoSOD3k}        & \multicolumn{4}{c}{Cosal2015}       \\ \cline{2-9} 
                        & $E_\xi$ & $S_m$ & $F_\beta$ & MAE   & $E_\xi$ & $S_m$ & $F_\beta$ & MAE   \\ \hline
Ours(Stage1)            & 0.850   & 0.815 & 0.792     & 0.088 & 0.912   & 0.870 & 0.860     & 0.065 \\
Ours(Stage2)            & 0.888   & 0.843 & 0.820     & 0.062 & 0.934   & 0.898 & 0.902     & 0.045 \\ \hline
\end{tabular}}
\label{tab_stage1}
\end{table}

{ \textbf{Training stages ablation studies.} In this paper, we introduce two stage training processes for the CoSOD task. The stage1 training process helps the network capture the regions tended to be salient, and the stage2 training process helps the network learn the co-salient regions. In the Fig.\ref{fig_stage1}, we show some examples where the stage1 model fails but the stage2 model succeeds. In the ``Apple Group" and ``Baseball Group", the stage1 model detects ``salient noise" which is suppressed in the stage1 model. In the ``Dumbbell Group" and ``Train Group", the object ``person" is detected in the stage1, and the real co-salient objects are omitted. Because in the SOD DUTS training dataset, the person is always the salient object and the other objects are the background. After fine-tuning the stage1 model on the sub-coco dataset~\cite{DBLP:conf/eccv/LinMBHPRDZ14}, the stage2 model would detect real co-salient objects. Quantitative comparison between the stage1 model and the stage2 model is in Table.\ref{tab_stage1}, and the stage2 model is obviously superior to the stage1 model. }

\section{Discussion} \label{discussion}

\begin{table}[t!]

\begin{minipage}{0.48\linewidth}
\centering
\caption{ { Results on Internet Dataset.}}
\scalebox{0.7}{
\begin{tabular}{c|cccc}
\hline
\multicolumn{1}{l|}{} & Airplane & Car   & Horse & $Avg.J$ \\ \hline
Li19                  & 0.830    & 0.930 & 0.760 & 0.840   \\
Li21                  & 0.840    & 0.920 & 0.830 & 0.863   \\
GCAGC                 & 0.835    & 0.919 & 0.809 & 0.854   \\
CADC                  & 0.833    & 0.916 & 0.806 & 0.851   \\
Ours                  & 0.868    & 0.951 & 0.802 & 0.874   \\ \hline
\end{tabular}}
\label{disc_internet}
\end{minipage}
\begin{minipage}{0.48\linewidth}\quad
\centering
\caption{ { Results on Pascal Dataset.}}
\scalebox{0.7}{
\begin{tabular}{c|cc}
\hline
\multicolumn{1}{l|}{} & $P$   & $J$   \\ \hline
Li19                  & 0.940 & 0.630 \\
Li21                  & 0.970 & 0.740 \\
GCAGC                 & 0.951 & 0.731 \\
CADC                  & 0.950 & 0.735 \\
Ours                  & 0.973 & 0.746 \\ \hline
\end{tabular}}
\label{disc_pascal}
\end{minipage}

\begin{minipage}{0.48\linewidth}
\centering
\caption{ { Results on iCoseg Dataset.}}
\scalebox{0.65}{
\begin{tabular}{c|ccccccccc}
\hline
\multicolumn{1}{l|}{} & bear2 & brownbear & cheetah & elephant & helicopter & hotballoon & panda1 & panda2 & $Avg.J$ \\ \hline
Li19                  & 0.901 & 0.897     & 0.920   & 0.902    & 0.760      & 0.917      & 0.902  & 0.898  & 0.887   \\
Li21                  & 0.928 & 0.941     & 0.891   & 0.916    & 0.852      & 0.951      & 0.948  & 0.921  & 0.921   \\
GCAGC                 & 0.921 & 0.930     & 0.916   & 0.911    & 0.845      & 0.941      & 0.941  & 0.895  & 0.913   \\
CADC                  & 0.922 & 0.931     & 0.918   & 0.908    & 0.840      & 0.938      & 0.939  & 0.891  & 0.911   \\
Ours                  & 0.933 & 0.934     & 0.926   & 0.932    & 0.883      & 0.964      & 0.949  & 0.902  & 0.927   \\ \hline
\end{tabular}}
\label{disc_icoseg}
\end{minipage}
\end{table}

\begin{figure*}[!hbpt]
\centering
\includegraphics[scale=0.5]{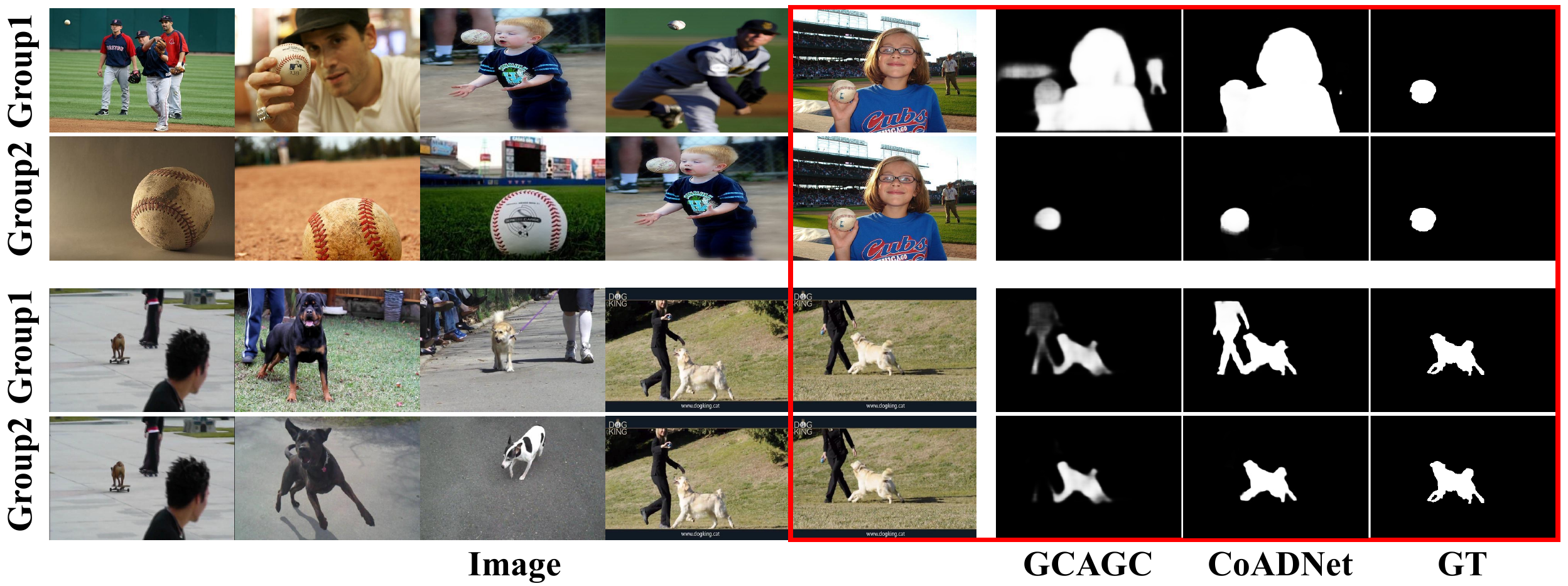}
\caption{The results generated from the methods GCAGC~\cite{DBLP:conf/cvpr/ZhangLSLC020} and CoADNet~\cite{DBLP:conf/nips/ZhangCHLZ20} in two different sub-groups.}
\label{fig_subgroup}
\end{figure*}

\begin{figure*}[!hbpt]
\centering
\includegraphics[scale=0.45]{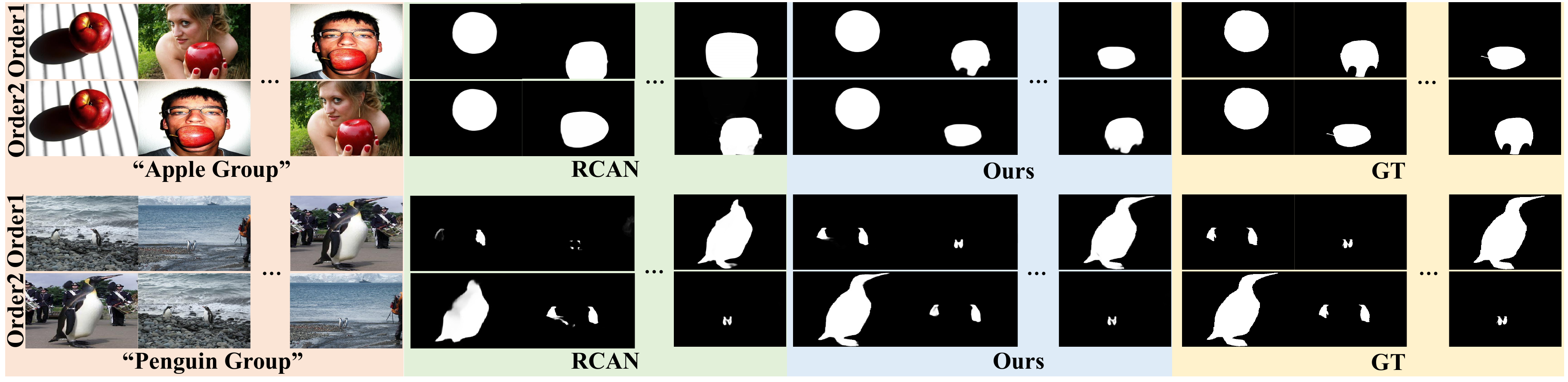}
\caption{The results generated from the method RCAN~\cite{DBLP:conf/ijcai/0061STSS19} and our proposed method in two different orders.}
\label{fig_subgroup2}
\end{figure*}

{

In this section, we make some further discussions of CoSOD/CoSEG task, including \textbf{the relations between CoSOD and CoSEG} and \textbf{more analysis of stability problem in the CoSOD/CoSEG task}. 

\textbf{The relations between CoSOD and CoSEG.} Both CoSOD and CoSEG aim at segmenting the co-occurring objects among an image group with unfixed image sizes, and network stability is an important factor for both two tasks, motivating us to use one RNN-based network to simultaneously address them. By the way, we find that the existing method~\cite{DBLP:journals/tip/TsaiLHQL19} also tries to design a unified network to address these two problems. To further verify this statement, we re-train the SOTA graph-based method GCAGC~\cite{DBLP:conf/cvpr/ZhangLSLC020} and CNN-based method CADC~\cite{DBLP:conf/iccv/ZhangHL021}, then apply them to the CoSEG task. The results are in Table.\ref{disc_internet}, Table.\ref{disc_pascal} and Table.\ref{disc_icoseg}. Compared to the CoSEG, CoSOD needs extra intra-saliency capturing module to let the network tend to focus on salient regions. GCAGC and CADC design the pertinent training mechanism, by altering the salient-based training datasets, to achieve this purpose. Herein, we only re-train these two methods on the sub-coco dataset~\cite{DBLP:conf/eccv/LinMBHPRDZ14}. It can be seen that these two models achieve competitive performance on the CoSEG benchmarks compared to other CoSEG models.

\begin{table}[!t]
\centering
\caption{ { The impact of different sub-groups.} }
\scalebox{0.8}{
\begin{tabular}{c|cccc}
\hline
\multirow{2}{*}{Models} & \multicolumn{4}{c}{Cosal2015}                                                  \\ \cline{2-5} 
                        & $E_\xi$           & $S_m$             & $F_\beta$         & MAE                \\ \hline
CoADNet                 & $0.910 \pm 0.022$ & $0.855 \pm 0.018$ & $0.850 \pm 0.022$ & $0.060 \pm 0.020$ \\
CoADNet(STD)            & 0.020             & 0.017             & 0.020             & 0.018              \\ \hline
GCAGC                   & $0.896 \pm 0.021$ & $0.839 \pm 0.020$ & $0.841 \pm 0.020$ & $0.080 \pm 0.019$  \\
GCAGC(STD)              & 0.019             & 0.019             & 0.017             & 0.018              \\ \hline
\end{tabular}}
\label{tab_subgroup}
\end{table}

\textbf{More analysis of stability problem in the CoSOD/CoSEG task.}
In this paper, we delve into the \textbf{stability} problem in the CoSOD/CoSEG task. In the most recent review article~\cite{9358006}, stability is highlighted as one of the most important issues currently unresolved in the CoSOD task. In fact, as an inherent problem, instability exists widely in CoSOD methods. Although many CNN-based methods~\cite{9358006,DBLP:conf/nips/ZhangCHLZ20,GLNet} and Graph-based methods~\cite{DBLP:conf/mm/0002JZT019,DBLP:conf/cvpr/ZhangLSLC020} have greatly advanced the development of CoSOD task in recent years, before our previous study~\cite{DBLP:conf/ijcai/0061STSS19}, the stability problem remained untouched. When dealing with image groups containing a variable number of images, these CNN-based methods and Graph-based methods detect co-salient objects by dividing the image group into image pairs or image sub-groups. Since there is no principle way of dividing image groups, this strategy inevitably makes the overall training as well as testing process unstable, which influences the application of the co-salient object detection. We conduct the experiment on CNN-based method CoADNet~\cite{DBLP:conf/nips/ZhangCHLZ20} and Graph-based method GCAGC~\cite{DBLP:conf/cvpr/ZhangLSLC020}. In the red boxes of Fig.\ref{fig_subgroup}, the same image results in the different detecting results when they are in the different sub-groups. Moreover, we repeatedly test these two methods in different sub-groups of Cosal2015 for 5 times, and the performance is shown in Table.\ref{tab_subgroup}. It can be seen that different sub-groups heavily affect the performance and we call this problem \textbf{sub-group instability}. To 
address the sub-group instability, in our previous work RCAN~\cite{DBLP:conf/ijcai/0061STSS19}, we propose the RNN-based framework to make use of all available information in an image group. However, as an RNN framework, when the images in the image group are assigned with different orders, RCAN faces another instability. We call this order-sensitivity problem \textbf{order instability}. As can be seen in the Fig.\ref{fig_subgroup2}, and the co-salient results of RCAN are variant under different orders. Hence, in this paper, we further explore how to alleviate the order instability of the RNN-based framework. In the Fig.\ref{fig_subgroup2}, when detecting the different orders, our proposed network can consistently achieve good results. Comparing the Table.\ref{tab_subgroup} to Table.\ref{Stability}, we find that introducing an RNN-based network already improves the CoSOD stability by addressing the sub-group instability. Finally, through the MSRU and COCL, the stability is further enhanced by a large margin, since the order instability is addressed. We believe that a sustained and in-depth exploration of stability issues is of great significance in advancing the CoSOD/CoSEG field. 
}

\section{Conclusion}
In this paper, we revisit the sequential modeling for CoSOD (CoSEG) task, then state the one drawback of existing models: order-sensitivity, which heavily affects the stability of proposed CoSOD (CoSEG) model. In this paper, inspired by RNN-based CoSOD (CoSEG) model, we first propose a multi-path stable recurrent unit (MSRU), containing dummy orders mechanisms (DOM) and recurrent unit (RU). Our proposed MSRU can not only help CoSOD (CoSEG) model capture robust inter-image relations, but also let the model have significant reduction in order-sensitivity, leading to a more stable training and inference process. Moreover, we design a cross-order contrastive loss (COCL) that can further improve order-sensitive problem by pulling close the feature embedding generated from different input orders. The performance on five widely used CoSOD datasets and three object co-segmentation datasets demonstrates the superiority of the proposed approach as compared to the SOTA methods. Through this paper, we investigate how RNNs can model orderless sequence tasks by using the COSOD (CoSEG) task as an example. When we design the mechanisms to address the order sensitivity problem, we find that RNNs are also capable of handling orderless sequence tasks. We hope that this paper can motivate future research for visual co-analysis tasks.
{\small
\bibliographystyle{IEEEtran}
\bibliography{IEEEfull}
}

\end{document}